\documentclass{article}
\usepackage{graphicx} 
\usepackage{amssymb}
\usepackage{amsmath}
\usepackage{hyperref}
\usepackage{booktabs}
\usepackage{iclr2024_conference}
\usepackage{url}
\usepackage{caption} 
\usepackage{wrapfig}
\usepackage{fancyhdr}
\newcommand\blfootnote[1]{%
  \begingroup
  \renewcommand\thefootnote{}\footnote{#1}%
  \addtocounter{footnote}{-1}%
  \endgroup
}

\title{Extreme Precipitation Nowcasting using Transformer-based Generative Models}
\author{
Cristian Meo\textsuperscript{1*} \And
Ankush Roy\textsuperscript{1*} \And 
Mircea Lică\textsuperscript{1*} \And 
Junzhe Yin\textsuperscript{1} \And
Zeineb Bou Cher\textsuperscript{1} \And 
Yanbo Wang\textsuperscript{1} \And
Ruben Imhoff\textsuperscript{2} \And
Remko Uijlenhoet\textsuperscript{1} \And 
Justin Dauwels\textsuperscript{1}
}

\date{January 2024}

\begin{document}
\setlength{\headheight}{22.54448pt}
\fancyhead[L]{Accepted for a spotlight talk at the Tackling Climate Change with Machine Learning workshop at ICLR 2024}
\maketitle
\blfootnote{$^*$Equal Contribution, $^1$Delft University of Technology, Netherlands $^2$ Deltares, Netherlands. Corresponding authors: \text{c.meo@tudelft.nl, A.Roy-8@student.tudelft.nl, M.T.Lica@student.tudelft.nl}.}
\vspace{-1 cm}
\begin{abstract}
This paper presents an innovative approach to extreme precipitation nowcasting by employing Transformer-based generative models, namely NowcastingGPT with Extreme Value Loss (EVL) regularization. Leveraging a comprehensive dataset from the Royal Netherlands Meteorological Institute (KNMI), our study focuses on predicting short-term precipitation with high accuracy. We introduce a novel method for computing EVL without assuming fixed extreme representations, addressing the limitations of current models in capturing extreme weather events. We present both qualitative and quantitative analyses, demonstrating the superior performance of the proposed NowcastingGPT-EVL in generating accurate precipitation forecasts, especially when dealing with extreme precipitation events. The code is available at \url{https://github.com/Cmeo97/NowcastingGPT}.
\end{abstract}

\section{Introduction}

The advent of climate change has escalated the frequency of intense rainfall events across various regions worldwide, leading to considerable societal and infrastructural impacts \citep{Alfieri2017GlobalPO, martinkova2020overview, Klocek2021MSnowcastingOP, Czibula2021AutoNowPAA, MalkinOndik2022Nowcasting}. Consequently, the ability to accurately forecast short-term shifts in rainfall patterns is gaining importance, attracting a growing body of research focus \citep{shi2015convolutional, trebing2021smaat, Luo2021PredRANNTS, Liu2022ASD}. The field of precipitation nowcasting, which involves predicting rainfall changes within a six-hour window, plays a crucial role in enabling timely responses to these rapid meteorological variations \citep{Veillette2020SEVIRA, MalkinOndik2022Nowcasting, Yang2022AATransUNetAA, prudden2020review}. 
In the context of escalating climate change impacts, the field of precipitation nowcasting is increasingly vital for mitigating the adverse effects of intense rainfall events. This research area empowers the development of advanced forecasting models that can provide accurate, short-term rainfall predictions. Such capabilities are essential for proactive disaster management and climate resilience strategies, enabling communities and infrastructure planners to prepare for and respond to extreme weather events more effectively, thereby contributing to meaningful efforts in addressing the climate crisis.
\section{Related Works}
Conventional nowcasting techniques, exemplified by frameworks such as PySTEPS \citep{pulkkinen2019pysteps}, adopt the ensemble-based methodology reminiscent of Numerical Weather Prediction (NWP) to incorporate uncertainty while modeling precipitation dynamics through the lens of the advection equation \citep{ravuri2021skilful}.  On the other hand, Deep learning-based approaches, leveraging extensive datasets of radar observations, can be trained without the constraints of predefined physical assumptions, significantly enhancing forecast accuracy \citep{ravuri2021skilful}.
In the last few years, precipitation nowcasting using deep learning models has been cast as a video prediction problem \citep{Haroan, Bai2022RainformerFE, Luo2021PredRANNTS, Liu2022ASD}, where given an input spatio-temporal sequence of $N$ frames $\boldsymbol{x}_{\text{in}} \in \mathbb{R}^{N \times H \times W \times C}$, where $H, W$ denote the spatial resolution and $C$ represents the image channels or the different type of measurements (e.g., radar maps, heat maps, etc), the goal is to predict the next $M$ frames $\boldsymbol{x}_{\text{out}} \in \mathbb{R}^{M \times H \times W \times C}$. Among the most notable advancements in the field, Generative Adversarial Networks (GAN) \cite{goodfellow2014generative} have emerged as a powerful approach, exemplified by methods such as DGMR \cite{ravuri2021skilful}, which employs both spatial and temporal discriminators to ensure the fidelity of generated sequences to the ground truth. Moreover, Transformer-based strategies \cite{vaswani2017attention} leverage an Autoregressive Transformer (AT) to model the hidden dynamics of precipitation maps \citep{preformer, Haroan}. For instance, \cite{Haroan} employs Nuwä \citep{wu2022nuwa}, an AT that uses a sparse attention mechanism, namely 3DNA \citep{wu2022nuwa}, to adeptly capture the complexities of precipitation dynamics. Moreover, \cite{Haroan} regularizes the hidden dynamics incorporating an Extreme Values Loss (EVL) to effectively model and predict extreme precipitation events, which are notoriously difficult to represent and predict. 
Although these models have improved in terms of prediction capabilities, they present critical drawbacks. Firstly, the prediction quality degrades very quickly, resulting in predicted sequences that are inconsistent over time. Secondly, the time required to generate the predicted sequences is extremely high, which is a critical problem considering that nowcasting predictions are supposed to predict the very next future. For instance, Nuwä+EVL \citep{Haroan} takes over 5 minutes to predict the next precipitation maps on a Nvidia RTX A6000. Furthermore, predicting and representing extreme precipitation events is still very challenging for all the proposed models. Although \cite{Haroan} uses an EVL as a regularizer, it assumes a predefined set of representations that should embed the extreme events features, assuming that the extreme features never change during training, which we believe to be a wrong inductive bias, since the topology of the hidden space changes during training. 
In this work, we propose NowcastingGPT, which follows VideoGPT framework \citep{yan2021videogpt}, employing a Vector Quantized-Variational AutoEncoder (VQ-VAE) \citep{van2017neural} to extract discrete tokens and an Autoregressive Transformer \citep{esser2021taming} to model the hidden dynamics. Moreover, we propose a novel approach to correctly compute the EVL regularization without assuming any fixed extreme representation. Moreover, we benchmark TECO \citep{yan2022temporally}, an efficient transformer-based video prediction model that generates temporally consistent frames, on the precipitation nowcasting task. Finally, we present both qualitative and quantitative comparisons of the considered models. 
\section{Methodology}
Video prediction tasks, at their core, involve forecasting the future frames of a video sequence based on past observations, akin to predicting the next scenes in a dynamic storyline. This challenge extends naturally to nowcasting, where the goal is predicting satellite imagery or radar maps, capturing the evolution of environmental and weather conditions over time. Both domains share the fundamental task of modeling and anticipating the progression of complex, time-varying patterns, making techniques developed for video prediction highly relevant and applicable to the realm of nowcasting. 

\subsection{Nowcasting as Video Prediction}
Video prediction tasks are known for their sample inefficiency, which poses significant challenges in learning accurate and reliable models. To address this, recent advancements have introduced spatio-temporal state space models, which typically consist of a feature extraction component coupled with a dynamics prediction module. These models aim to understand and predict the evolution of video frames by capturing both spatial and temporal relationships. Notable examples include Nuwä \citep{wu2022nuwa} and VideoGPT \citep{yan2021videogpt} which, leveraging the space-efficient VQ-VAE feature extraction, and the powerful sequence modeling capabilities of Autoregressive Transformers, can achieve a deeper understanding of the underlying video dynamics, leading to more accurate predictions of future frames. We define the video prediction backbone of the proposed nowcasting model following the VideoGPT framework, using a VQ-VAE as a feature extractor and an Autoregressive Transformer \citep{esser2021taming} to learn the latent space dynamics and predict the future precipitation maps. A detailed description of the NowcastingGPT model employed in this work can be found in appendix \ref{appendix:VideoGPT}.
\subsection{Extreme Value Loss Regularization}
When dealing with imbalanced data, the standard cross-entropy loss often falls short,  particularly when classifying extreme events. To address this, the Extreme Value Loss (EVL) has been introduced as a more effective alternative, designed to balance the disparities between extreme and non-extreme cases in time series data \cite{evtTimeseries}:
\begin{equation}
\text{EVL}(u_{t}, v_{t}) = -\beta_{1}\left[1-\frac{u_{t}}{\gamma}\right]^{\gamma} v_{t} \log(u_{t}) -\beta_{0}\left[1-\frac{1-u_{t}}{\gamma}\right]^{\gamma}(1-v_{t}) \log(1-u_{t}),
\label{eqtn:EVL}
\end{equation}
\begin{wrapfigure}{r}{0.5\textwidth}
    \centering
    \includegraphics[width=0.8\linewidth,trim={0.0cm 0 0.0cm 0},clip]{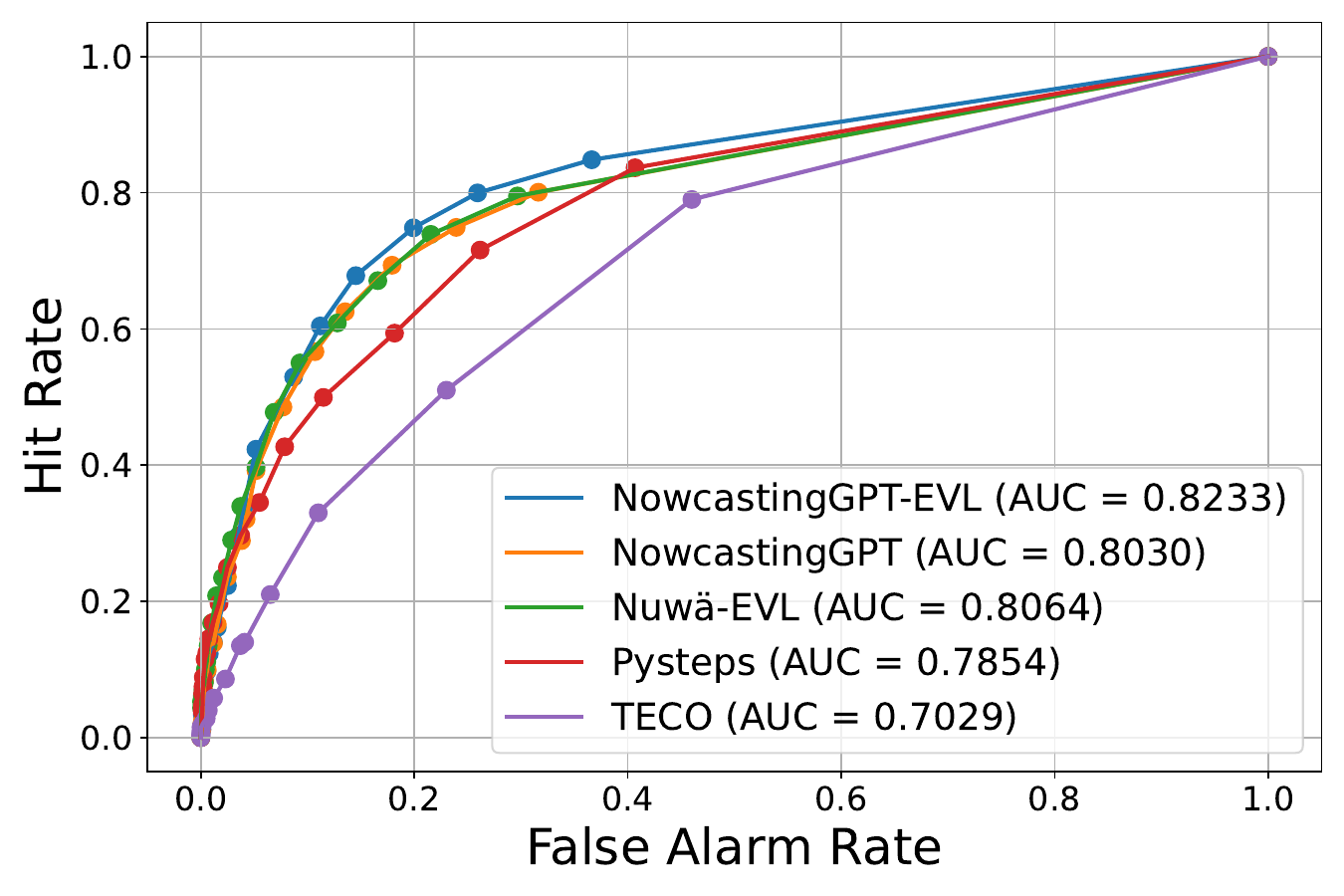}
    \caption{ROC Curve for extreme event detection. Thresholds between $0.5$ and $10$ for precipitation values are used to define an extreme event. NowcastingGPT-EVL had the highest AUC, outperforming all other baselines.}
    \label{img:roccurve}
\end{wrapfigure}
where $v_{t}$ represents the ground truth labels (e.g., extreme/not extreme), $u_{t}$ the predicted probabilities, and $\gamma$, a hyperparameter of the Generalized Extreme Value (GEV) distribution. By incorporating $\beta_{0}$ and $\beta_{1}$, which reflect the proportions of non-extreme and extreme tokens, EVL effectively balances the learning process. When regularizing an Autoregressive Transformer the EVL  enhances the model's ability to predict and represent extreme events. To this end, we define a classifier that dynamically predicts extreme labels. As a result, we can use the EVL to regularize the Autoregressive Transformer learning behavior and improve its ability to capture extreme phenomena in data sequences. A detailed description of the classifier can be found in appendix \ref{appendix:Classifier}, while the full derivation of the EVL loss can be found in appendix \ref{appendix:EVL}.
\section{Experiments}
In this section, we design empirical experiments to understand the performance of NowcastingGPT-EVL and its
potential limitations by exploring the following questions: 
(1) Does EVL regularization improve the nowcasting performances of the proposed model? (2) How does time consistency affect downstream results? (3)  Does learning extreme representations provide a more effective inductive bias compared to relying on predefined ones?
\subsection{Dataset and Experimental setup}
\label{exp:dataset}
Our nowcasting study aims to predict precipitation patterns up to three hours into the future. This approach generates a series of six future precipitation maps, each separated by 30 minutes, conditioned on three previous precipitation maps used as input. Specifically, we use radar maps defined  $256\times256$ images, which  include a vast expanse of the national territory and all critical catchment areas, following the approach used in \citep{Haroan}. More details about the dataset are described in appendix \ref{appendix:dataset}.
We compare the proposed model to a classic benchmark, namely Pysteps \citep{pulkkinen2019pysteps}, a temporally consistent video prediction benchmark, TECO \citep{yan2022temporally}, the NowcastingGPT model which is described in the appendix \ref{nowcastinggpt} and Nuwä-EVL proposed by \cite{Haroan}, which uses fixed latents to represents extreme features.  An in-depth description of the considered baselines can be found in appendix \ref{appendix:Baselines}. To quantitatively assess the experiments we use visual fidelity metrics, such as Mean Squared Error (MSE), Mean Absolute Error (MAE) and Pearson Correlation Score (PCC), and nowcasting metrics, such as Critical Success Index (CSI), False Alarm Rate (FAR) and Fractional Skill Score (FSS). Since fidelity metrics cannot capture extreme event classification, we plot an ROC curve of the extremes to assess the considered baselines in terms of extreme classification capabilities. A detailed description of these metrics can be found in appendix \ref{appendix:Metrics}
\subsection{Experimental Results}
\label{experimental}
We test the performance of the proposed models by using the extreme precipitation test set described in appendix \ref{appendix:dataset}. Table \ref{table:results} showcases the effectiveness of these methods against a series of metrics that asses the quality and validity of the predictions. The proposed NowcastingGPT-EVL outperforms the other models on the majority of metrics and close second on the rest. The ROC curve in Figure \ref{img:roccurve} demonstrates how NowcastingGPT-EVL outperforms all other methods on extreme event detection at different thresholds. Figure \ref{img:fig3} illustrates the predicted maps of all considered baselines. While NowcastingGPT presents meaningful predictions over all time steps, Nuwä-EVL deteriorates substantially. Indeed, we believe that when Nuwä-EVL extreme representations get updated by the AT, the VQ-VAE is not able to recognize the extreme latents anymore, which by design are supposed to be fixed, predicting images that do not resemble the ground truth maps semantics. Remarkably, the graphs presented in Appendix \ref{appendix:results} demonstrate that TECO achieves results on par with other methods, despite having fewer parameters and a more efficient sampling time, and exhibits superior temporal consistency compared to alternative approaches.

\begin{table}
\caption{Quantitative results of the proposed methods. Each value represents the average and standard deviation over the means and standard deviations of each of the $6$ lead times. The description for each metric can be found in appendix \ref{appendix:Metrics}. For statistically meaningful results, we consider $3$ different seeds for each entry.}
\centering
\scalebox{0.75}{
\begin{tabular}{cccccccccc} \toprule
    {$ $} & {Nuwä-EVL} & {NowcastingGPT} & {PySTEPS} & {TECO} & {NowcastingGPT-EVL} \\ \midrule

    PCC $(\uparrow)$ & $0.15$ & $\underline{0.20} \pm 0.002$ & $0.14$ & $0.10\pm 0.002$ & $\textbf{0.22} \pm 0.002$ \\ \midrule
    MSE $(\downarrow)$  & $4.85 $ & $\underline{3.60} \pm 0.02$ & $6.22 $ & $3.65 \pm 0.008$ & $\textbf{3.45} \pm 0.02$ \\ \midrule
    MAE $(\downarrow)$ & $1.00 $  & $0.72 \pm 0.005$ & $0.93 $ & $\textbf{0.68} \pm 0.001$& $\underline{0.69} \pm 0.005$\\ \midrule
    CSI(1mm) $(\uparrow)$  & $\textbf{0.23} $  & $0.21 \pm 0.002$ & $0.21 $ & $0.07 \pm 0.001$ & $\underline{0.22} \pm 0.002$ \\ \midrule
    CSI(2mm) $(\uparrow)$  & $\textbf{0.13}$   & $0.11 \pm 0.001$ & $ \underline{0.12} $ & $0.03 \pm 0.001$ & $\underline{0.12} \pm 0.001$ \\ \midrule
    CSI(8mm) $(\uparrow)$   & $0.008$   & $0.005 \pm 0.0005$ & $\textbf{0.01} $ & $0.001 \pm 0.0009$ & $\underline{0.009} \pm 0.0005$ \\ \midrule
    FAR(1mm) $(\downarrow)$  & $ 0.61 $   & $0.59 \pm 0.002$ & $ \textbf{0.55} $ & $0.69 \pm 0.002$ & $\underline{0.59} \pm 0.002$ \\ \midrule
    FAR(2mm) $(\downarrow)$  & $ 0.76 $  & $0.71 \pm 0.0007$ & $ \textbf{0.70} $ & $0.78 \pm 0.004$ & $\underline{0.71} \pm 0.0007$ \\ \midrule 
    FAR(8mm) $(\downarrow)$  & $ 0.85 $  & $0.59 \pm 0.003$ & $ 0.89 $ & $\textbf{0.49} \pm 0.006$ & $\underline{0.52} \pm 0.003$ \\ \midrule
    FSS(1km) $(\uparrow)$  &  $ 0.35 $   & $0.49 \pm 0.003$ & $ 0.32 $ & $\underline{0.49} \pm 0.003$ & $\textbf{0.52} \pm 0.003$ \\ \midrule
    FSS(10km) $(\uparrow)$  & $ 0.42 $   & $\underline{0.55} \pm 0.004$ & $ 0.41 $ & $0.46 \pm 0.003$ & $\textbf{0.58} \pm 0.004$  \\ \midrule
    FSS(20km) $(\uparrow)$  & $ 0.48 $   & $\underline{0.59} \pm 0.004$ & $ 0.47 $ & $0.42 \pm 0.003$ & $\textbf{0.62} \pm 0.004$  \\ \midrule
    FSS(30km) $(\uparrow)$  & $0.52 $  & $\underline{0.62} \pm 0.004$ & $ 0.51 $ & $0.37 \pm 0.002$ & $ \textbf{0.65} \pm 0.004$ \\\bottomrule
\end{tabular}
}
\label{table:results}
\end{table}

\section{Conclusion \& Discussion}
This work proposes NowcastingGPT-EVL, a video prediction model regularized using an EVL regularizer, validating the efficacy of using EVL for nowcasting extreme precipitation events. Our findings reveal that the proposed model outperforms existing methods in various downstream metrics, providing more accurate predictions. The study highlights the importance of addressing data imbalances and the dynamic nature of extreme events in model training. As future work, we aim to assess the prediction capabilities of the different models on an existing and widely used benchmark dataset (e.g., SEVIR \citep{SEVIRDataset}). The successful application of NowcastingGPT-EVL underscores the potential of Transformer-based models in enhancing predictive capabilities for critical meteorological forecasting tasks, paving the way for future advancements in the field.

\bibliography{iclr2024_conference}
\bibliographystyle{iclr2024_conference}

\newpage
\section{Appendix}\label{appendix:VideoGPT}


\subsection{Dataset}
\label{appendix:dataset}
The reflectivity measurements in the KNMI \citep{sluiter2012interpolation} dataset allows for the estimation of rainfall rates through the application of a Z-R transformation, enabling a nuanced river catchment-level analysis to evaluate the model's effectiveness in real-world scenarios. Ideally, extreme rainfall events are identified based on the distribution of the highest annual rainfall amounts. However, given the limited span of our dataset, which encompasses only 14 years, the dataset provides an insufficient quantity of annual maximums for effective model training and evaluation. Consequently, we have broadened the criteria for what constitutes extreme rainfall. Within this study, an event is classified as extreme if the average precipitation over a three-hour period within a catchment area ranks within the top 1\% of all observations recorded from 2008 to 2021. This adjustment allows for a more feasible and statistically sound basis for distinguishing significant precipitation events during the study period. The training dataset consists of 30632 sequences of images with each sequence consisting of 9 images (T-60, T-30, T, T+30, T+60, T+90, T+120, T+150, T+180 minutes) spanning from 2008-2014. The validation dataset consists of 3560 sequences of images with the same sequence length from year 2015-2018. The testing dataset utilised to evaluate the performance of the different models in this study, consists of 357 nationwide extreme events from 2019-2021, corresponding to 3927 events in the catchment regions.

\subsection{Metrics}
\label{appendix:Metrics}
The effectiveness of any predictive model is critically assessed through objective metrics that encapsulate its performance capabilities. In our endeavor to evaluate the impact of integrating the EVL regularization, we utilize a comprehensive set of performance metrics:

- \textit{Mean Absolute Error (MAE)}: MAE quantifies the average magnitude of errors in the predictions. It's computed as the mean of the absolute differences between the predicted values and the actual observations, offering a clear and intuitive metric for prediction accuracy.

- \textit{Mean Squared Error (MSE)}: MSE measures the average of the squares of the errors between the predicted and actual values, providing a more sensitive metric that penalizes larger errors more severely than MAE.

- \textit{Pearson Correlation Coefficient (PCC)}: PCC assesses the linear correlation between the predicted and observed datasets, yielding a value between -1 and 1, where 1 indicates perfect positive correlation, -1 indicates perfect negative correlation, and 0 signifies no linear correlation.

- \textit{Critical Success Index (CSI)}: CSI is utilized to evaluate the precision of forecasted events, particularly the successful prediction of specific events. This study examines CSI at two distinct precipitation thresholds: 1mm for light precipitation and 8mm for heavy precipitation, thus catering to varying intensities of rainfall.

- \textit{False Alarm Rate (FAR)}: FAR is calculated as the proportion of false positive predictions relative to the total number of positive forecasts (false positives plus true positives), offering insight into the model's tendency to incorrectly predict events that do not occur.

- \textit{Fractional Skill Score (FSS)}: FSS measures the model's forecast accuracy at specific spatial scales, facilitating an understanding of how well the model performs both locally and over broader areas. In this study, FSS is evaluated at 1km, 10km, 20km and 30km scales to discern the model's effectiveness at varying geographical extents.

While both MAE and MSE loss quantify the quality of the predictions, they are not able to capture the model's capability to detect extreme events. Thus, we make use of a Receiver Operating Characteristic (ROC) curve to asses hit rate detection of extreme precipitation events. The curve is constructed using a set of thresholds that are used to define an event.

\subsection{Baselines Comparison}
\label{appendix:Baselines}
Motivated by the overall inefficiency in nowcasting methods, we consider TECO \citep{yan2022temporally} as a point of reference in benchmarking both training and sampling time of Transformer-based nowcasting models. TECO aims to increase sampling efficiency by replacing the common autoregressive prior with a masked token prediction objective, introduced by \cite{chang2022maskgit}. Using the discrete tokens from a VQ-VAE, the model learns to predict a randomly generated mask sampled at each timestamp, allowing for orders of magnitude improvement in sampling speed. Moreover, TECO manages to drastically decrease training time by using DropLoss, a trick that allows the model to consider only a subset of the frames that compose the video. Moreover, to be consistent with the literature, we consider 
PySTEPS \citep{pulkkinen2019pysteps}, a widely used numerical model for short-term precipitation predictions that achieves remarkable results in nowcasting \citep{imhoff2020spatial}. In appendix \ref{appendix:Baselines} table \ref{table:eff} presents a quantitative comparison between all proposed methods in terms of number of parameters, training and generation efficiency. 

Interestingly, TECO, showcases orders of magnitude more efficient generation time and cuts the training time by approximately 100 hours compared to its closest counterpart. Furthermore, with a generation time of $322.86$ seconds, Nuwä-EVL constitutes a good indicator for the sampling efficiency of autoregressive models. 


\begin{table}
\centering
\caption{Comparison of the proposed methods in terms of number of parameters, training time and generation time. Generation time refers to the time required on average to sample a sequence from the dataset defined in Section \ref{appendix:dataset}. Training time is computed in terms of GPU hours.}
\scalebox{0.80}{
\begin{tabular}{cccccccccc} \toprule
    {$ $} & {Nuwä-EVL} & {NowcastingGPT} & {PySTEPS} & {TECO} & {NowcastingGPT-EVL} \\ \midrule
    Number of parameters  & $772,832$ M & $402,735$ M & - & $165,960$ M & $520,374$ M \\ \midrule
    Training time  & $672$h  & $240$h & - & $155$h & $264$h \\ \midrule
    Generation time  & $322.86$s & $38.90$s & $9.34$s & $0.51$s & $43.10$s \\\bottomrule

\end{tabular}
}
\label{table:eff}

\end{table}
\subsection{NowcastingGPT-EVL Description}

\begin{figure}[ht]
    \centering
    \includegraphics[width=8cm]{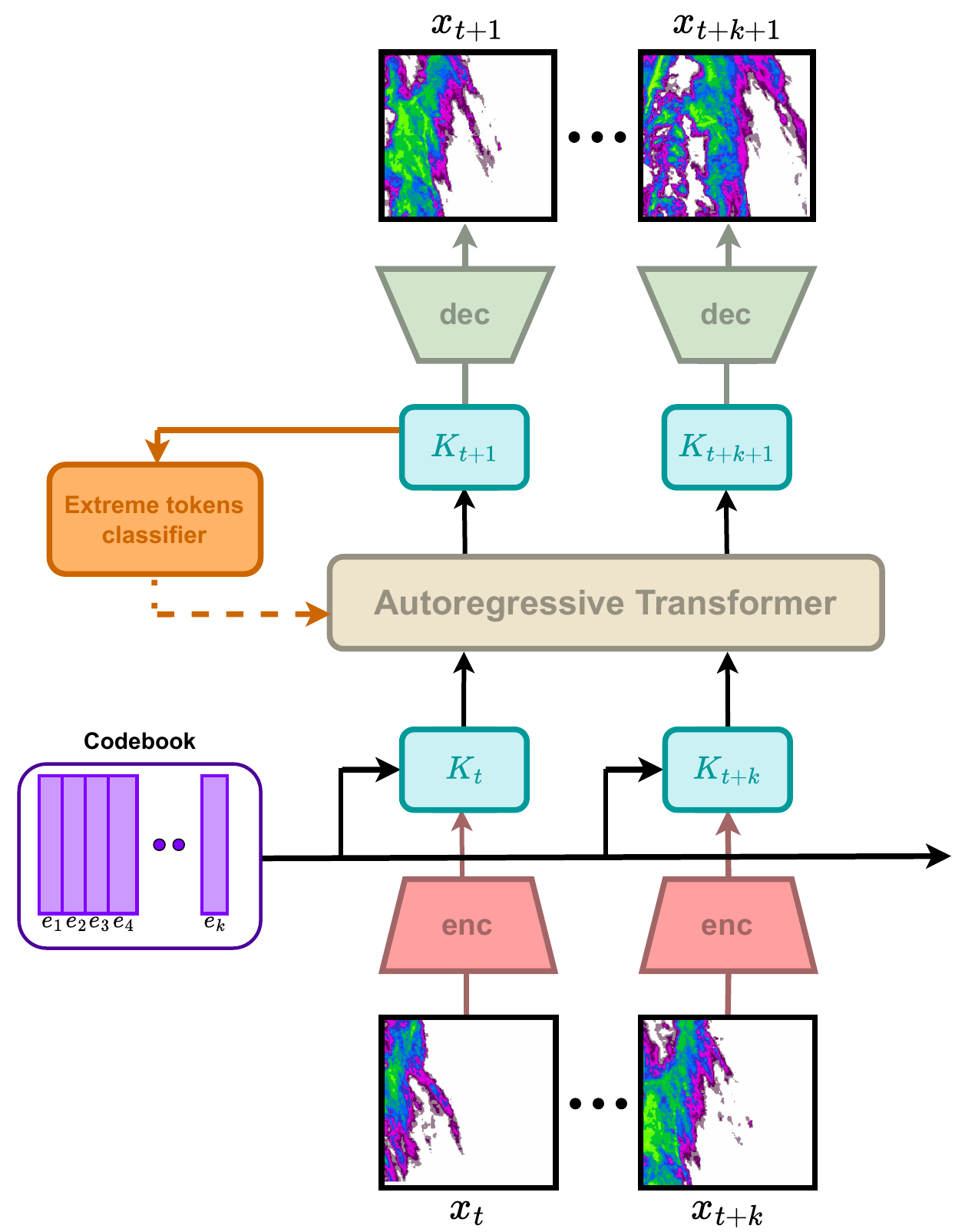}
    \caption{The image shows the NowcastingGPT-EVL model architecture. The VQ-VAE Encoder and Decoder are depicted in red and green respectively. The Extreme tokens classifier is depicted in orange, it takes the predicted tokens as input from the transformer and outputs the probabilities $u_t$ used in the EVL loss. The dashed line indicates that the output of the Classifier is only used to optimize the transformer and not as input. }
    \label{img:VideoGPT-EVL}
\end{figure}

In this section, we describe the used model. Figure \ref{img:VideoGPT-EVL} illustrates the model architecture. The following subsections describe the three main components: VQ-VAE, Autoregressive Transformer, and Extreme tokens classifier.

\subsubsection{Vector Quantized Variational Autoencoder}

The Vector Quantized Variational AutoEncoder (VQVAE) \citep{van2017neural} introduces a novel approach by utilizing Vector Quantization to encode inputs into discrete latent representations, moving away from continuous feature representations. This method is effective in capturing the complex, multi-dimensional features of data. VQVAE operates on an encoder-decoder framework with a discrete codebook, where the encoder compresses input data into a discrete set of codes, preserving essential features through a reduction in spatial dimensions and an increase in feature channels. The decoder then reconstructs the input from these codes, aiming for a close approximation to the original, thereby enabling efficient and structured data representation suitable for tasks like image reconstruction. The encoder consists of $5$ downsampling layers each containing $2$ ResNet blocks, thus reducing the spatial dimension of the input to the following resolutions: $128 \rightarrow 64 \rightarrow 32 \rightarrow 16 \rightarrow 8$. Furthermore, the last stage of the encoder includes an attention block used to capture the relationships between features before the quantization step. In order to obtain the reconstructed image from the discrete codes, we use a decoder that mirrors the structure of the encoder.


To facilitate the training of the VQVAE model, a set of distinct loss functions are harnessed and subjected to optimization. These loss functions encompass the reconstruction loss, the commitment loss, and the perceptual loss. 
\begin{equation}
\begin{aligned}
\mathcal{L}(E, D, \mathcal{Z})=\|x-\hat{x}\|_2^2 & +\left\|\operatorname{sg}[E(x)]-z_{\mathbf{q}}\right\|_2^2 
& +\left\|\operatorname{sg}\left[z_{\mathbf{q}}\right]-E(x)\right\|_2^2 & +\mathcal{L}_{\text {perceptual}}(x, \hat x),
\end{aligned}
\label{eq:VQVAE}
\end{equation}
\\
where $\hat{z}= E(x)\in \mathbb{R}^{h \times w \times n_z} $ represents the encoded image while $\hat{x} = D(z_q)$ is the reconstructed image using $z_{\mathbf{q}}$. We obtain $z_{\mathbf{q}}$ using an element-wise quantization $q(.)$ of each spatial code $\hat{z}_{ij} \in \mathbb{R}^{n_z} $ given by

$$
z_{\mathbf{q}}=\mathbf{q}(\hat{z}):=\left(\underset{z_k \in \mathcal{Z}}{\arg \min }\left\|\hat{z}_{i j}-z_k\right\|\right) \in \mathbb{R}^{h \times w \times n_z}.
$$

\subsubsection{Autoregressive Transformer}
\label{nowcastinggpt}


In order to model the dynamics between consecutive precipitation maps, we use an Autoregressive Transformer. For training the model, we utilize the ground truth precipitation maps which are quantized into $\mathbf{z}_q = q(\mathbf{E}(x))$, generating a sequence  $\mathbf{s} \in \{0, \ldots, |\mathbf{Z}|-1\}^{h \times w}$, corresponding to the respective indices of the VQVAE codebook. Subsequently, these indices are transformed into continuous vector representations using an embedder. In order to provide sequence order information to the Transformer, the representations are augmented with positional embeddings. These are then processed by the Transformer, which outputs the logits generated by the head module. These logits represent the probability of using a specific token and are used to compute a cross-entropy loss. The loss compares the predicted probabilities given by the model with the actual token probabilities:
\begin{equation}
\label{eq:CEPLoss}
\mathcal{L}_{\text{Transformer}} = \mathbb{E}_{x\sim p(x)}[-\log \prod_{i=1}^{N} p(\mathbf{s}_i | \mathbf{s}_{<i})] 
\end{equation}
which, given a sequence of indices $ \mathbf{s}_{<i}$, the Transformer is trained to predict the distribution of the consecutive indices $\mathbf{s}_i$. The AT employs a causal attention mechanism, where the non-causal entries of $QK^T$, those below the diagonal of the attention matrix, are set to $-\infty$. As a result, the attention mechanism accesses only previously seen or current tokens when predicting the next one in a sequence, enabling efficient and context-sensitive output production. We use the architecture described above to define our ablation model NowcastingGPT.

The Autoregressive Transformer for NowcastingGPT-EVL has the EVL loss function incorporated in it so the overall loss function for the AR transformer is given as:
\begin{equation}
\mathcal{L}_{\text{Transformer(NowcastingGPT-EVL)}} = \mathcal{L}_{\text{Transformer}} + \lambda [\text{EVL}(u_{t}, v_{t})].
\label{ART-EVL}
\end{equation}
The value of $\lambda$ in the equation above is chosen as 0.5. 
\subsubsection{Binary Classifier}
\label{appendix:Classifier}
For the classification of the tokens into extreme or non-extreme, a transformer is incorporated along with the auto-regressive transformer. The input to this transformer are the sequence of tokens that are generated from the auto-regressive transformer during its training phase. The model has 6 layers, 1024 embedding dimension and, a  total number of 8 heads. The transformer is trained using a standard binary cross entropy loss function where, the ground truth labels $v_t$ are calculated on the basis of averaged precipitation over a threshold of 5mm. In this way, all the tokens corresponding to an extreme/non-extreme event get classified along with the training of the auto-regressive transformer. The classifier generates logits for the two classes (extreme and, non-extreme) which are then passed through a softmax layer to generate probabilities. These probabilities act as the input to the EVL loss function mentioned in equation (\ref{eqtn:EVL}) for the term $u_t$. The values for $\beta_{0}$ and $\beta_{1}$ are taken as 0.95 and 0.05 respectively since, top 5\% of the events are considered as extreme events.
The value of $\gamma$ for EVL was set to 1, as this setting demonstrated optimal performance.

\subsection{Mathematical Proof of the EVL loss function}
\label{appendix:EVL}
As mentioned in \cite{Coles2001}, if there is a sequence of independent and identically distributed (i.i.d)
random variables as $ X_1, X_2, \ldots, X_n $, having marginal distribution function $F$. It is natural to regard as extreme events those of the $X_i$ that exceed some high threshold $u$. Denoting an arbitrary term in the $X_i$ sequence by $X$, it follows that
a description of the stochastic behavior of extreme events is given by the
conditional probability:
\begin{equation}
\operatorname{Pr}\{X>u+y \mid X>u\}=\frac{1-F(u+y)}{1-F(u)}, \quad y>0.
\label{eqtn 1}
\end{equation}
Starting from the L.H.S we have:
$$
\begin{aligned}
\operatorname{Pr}\{X>u+y \mid X>u\},
\end{aligned}
$$
and using the formula $P(x \mid y)=\frac{P(x, y)} {P(y)}$:
$$
\begin{aligned}
\operatorname{Pr}\{X>u+y \mid X>u\} & =\frac{P(X>u+y, X>u)}{P(X>u)}  \\
& =\frac{P(X>u+y)}{P(X>u)}.
\end{aligned}
$$
Applying the formula, $P(X>x) = 1-F(x)$ we get,
$$
=\frac{1-F(u+y)}{1-F(u)}.
$$
If the parent distribution $F$ was known to us then the distribution of threshold exceedances in equation \ref{eqtn 1} would also be known, however, that is not the case. \cite{Coles2001} suggests the application of Extreme Value Theory (EVT) for the approximation of the distribution of maxima of long sequences when the parent population function (distribution) $F$ is unknown. For the sequence of R.Vs mentioned above (with common distribution function $F$), we use maximum order statistics to characterize extremes :
\begin{equation}
M_n=\max \left\{X_1, X_2, X_3, \ldots X_n\right\}, \stackrel{P}{\rightarrow} x^*, n \rightarrow \infty .
\end{equation}
where $\stackrel{P}{\rightarrow}$ denotes convergence in probability and, $x^*$ denotes the right end point which is $x^*=\sup \{x: F(x)<1\}$
Therefore, for a large $n$ we have :
\begin{equation}
P\left(\max \left(X_1, X_2, \ldots, X_n\right) \leqslant x\right)=Pr\left(X_1 \leqslant x, X_2 \leqslant x, X_3 \leqslant x, \ldots X_n \leqslant x\right),
\label{eqtn GEV}
\end{equation}
since, they are i.i.d we can also write equation \ref{eqtn GEV} as,
$$
\begin{aligned}
P\left(\max \left(X_1, X_2, \ldots, X_n\right) \leqslant x\right)=[\operatorname{Pr}(X \leqslant x)]^n
& = [\operatorname{F}(x)]^n.
\end{aligned}
$$
Hence, from $$
\begin{aligned}
& [F(x)]^n \rightarrow 0 \text { for } x < x^* \\ 
& [F(x)]^n \rightarrow 1 \text { for } x \geqslant x^*,
\end{aligned}
$$
it can be said said that $[F(x)]^n$ is a degenerate function as it converges to a single point when $n$ becomes sufficiently large. To mitigate this, EVT suggests that for a sequence of constants $a_n > 0$ and real $b_n$ there is a non-degenerate distribution function $G$ stated as:
\begin{equation}
\lim _{n \rightarrow \infty}\left[F\left(a_n x+b_n\right)\right]^n=G(x),
\label{Gev_def}
\end{equation}
where $G(x)$ is the Generalised Extreme Value distribution function (GEV). The GEV is given by:
\begin{equation}
G(x)=\exp \left\{-\left[1+\xi\left(\frac{x-\mu}{\sigma}\right)\right]^{-1 / \xi}\right\},
\label{GEV function}
\end{equation}
where $\mu$ is the location parameter, $\sigma$ is the scale parameter and $\xi$ is the shape parameter. Also, equation (\ref{Gev_def}) can be written as:
$$
\begin{aligned}
& \left[F\left(a_n x+b_n\right)\right]^n \approx G(x) \\
\implies & \left[F\left(x\right)\right]^n \approx G\left\{\left(x-b_n\right) / a_n\right\} \\
\implies & \left[F\left(x\right)\right]^n =G^*(x),
\end{aligned}
$$
where $G^*$ is another member of the GEV family.  
\cite{Coles2001} mentions that if equation (\ref{Gev_def}) allows the approximation of $[F\left(a_n x+b_n\right)]^n$ by a member of the GEV family for large $n$, then $[F\left( x\right)]^n$ can also be approximated using a different member of the GEV family ($G^*(x))$ which has the same definition as mentioned in \ref{GEV function} but with different values of $\mu$, $\sigma$ and $\xi$.
Therefore, we can then write :
\begin{equation}
[F(x)]^n \approx \exp \left\{-\left[1+\xi\left(\frac{x-\mu}{\sigma}\right)\right]^{-1 / \xi}\right\}.
\end{equation}
Taking natural logarithm on both sides,
$$
n(\ln F(x)) \approx\ -\left[1+\xi\left(\frac{x-\mu}{\sigma}\right)\right]^{-1 / \xi}. \
$$
For large values of $x$, a Taylor expansion implies that,
$$
\ln F(x) \approx-\{1-F(x)\}.
$$
Substituting this in the above equation we get,
\begin{equation}
1-F(x) \approx \frac{1}{n}\left[1+\xi\left(\frac{x-\mu}{\sigma}\right)\right]^{-1 / \xi}.
\label{Taylor expand}
\end{equation}
Now, we substitute the above result obtained in the R.H.S of equation (\ref{eqtn 1}) for a large $u$ and $y>0$ as,
$$
1-F(u) \approx \frac{1}{n}\left[1+\xi\left(\frac{u-\mu}{\sigma}\right)\right]^{-1 / \xi}
$$
and,
$$
1-F(u+y) \approx \frac{1}{n}\left[1+\xi\left(\frac{u+y-\mu}{\sigma}\right)\right]^{-1 / \xi}.
$$
Therefore, we can write equation (\ref{eqtn 1}) as:
\begin{equation}
\begin{aligned}
\operatorname{Pr}\{X>u+y \mid X>u\} & \approx \frac{n^{-1}[1+\xi(u+y-\mu) / \sigma]^{-1 / \xi}}{n^{-1}[1+\xi(u-\mu) / \sigma]^{-1 / \xi}} \\
& =\left[1+\frac{\xi y / \sigma}{1+\xi(u-\mu) / \sigma}\right]^{-1 / \xi} \\
& =\left[1+\frac{\xi y}{\tilde{\sigma}}\right]^{-1 / \xi},
\end{aligned}
\label{GPD approximation}
\end{equation}
where $\tilde{\sigma} = \sigma + \xi(u - \mu)$.
This distribution function is known as the Generalised Pareto Distribution (GPD) function which helps in modeling observations over a large enough threshold $u$ (Peaks Over Threshold method - POT) and is written formally as :
\begin{equation}
H(y)=1-\left(1+\frac{\xi y}{\tilde{\sigma}}\right)^{-1 / \xi},
\label{GPD}
\end{equation}
defined on $\{y: y>0$ and $(1+\xi y / \tilde{\sigma})>0\}$, where
$$
\tilde{\sigma}=\sigma+\xi(u-\mu) .
$$
According to \cite{Coles2001}, the above relation in equation \ref{GPD approximation} implies that, if block maxima have approximating distribution $G$, then threshold excesses have a corresponding approximate distribution within the GPD family ($H$). Also, the parameters of GPD can be uniquely determined by those of the associated GEV distribution of block maxima. Moreover, the GEV distribution function and the GPD distribution function are related to each other since they have the same shape parameter $\xi$ so we can derive a rough mathematical relation between these two distribution functions as:
\begin{equation}
H(y) = 1 + \ln(G(y)),
\label{Rel. GPD & GEV}
\end{equation}
for some location $(\mu)$ and shape $(\sigma,\tilde{\sigma})$ parameters. This relationship is also mentioned in the paper \cite{Gencay2004-gb} which utilises EVT and Value-at-Risk for relative performance of stock market returns in emerging markets.
We can rewrite equation (\ref{eqtn 1}) with the help of the derived results in equations (\ref{GPD approximation}) and (\ref{GPD}) as:
\begin{equation}
\begin{aligned}
\frac{1-F(u+y)}{1-F(u)}=\left[1+\frac{\xi y}{\tilde{\sigma}}\right]^{-1 / \xi} 
& \implies \frac{1-F(u+y)}{1-F(u)}=1-H(y) \\
& \implies 1-F(u+y) \approx (1-F(u))(1-H(y)). \\
\label{Main tail approx.}
\end{aligned}
\end{equation}
This equation is the main equation for the tail approximation of observations exceeding a threshold $u$ and matches with the tail approximation equation mentioned in \cite{De_Haan2006-eu} as:
\begin{equation}
1-F(x) \approx(1-F(t))\left\{1-H_\xi\left(\frac{x-t}{f(t)}\right)\right\}, x>t,
\label{DeHaan Tail approx.}
\end{equation}
where $H_\xi$ is the GPD function with the shape parameter $\xi$.
Therefore, we use the result derived in equation (\ref{Main tail approx.}) to derive the weights of the EVL loss function mentioned in the paper  \cite{Haroan}. However, the authors utilise the GEV distribution function to define the underlying distribution of the time series data used in the paper. The goal of the paper is to predict outputs $Y_{T: T+K}$ in the future given the observations $(X_{1: T}, Y_{1: T}$ and future inputs $X_{T: T+K}$. For the sake of convenience, the authors define $X_{1: T} = \left[x_1, \cdots, x_T\right]$ and $Y_{1: T}=\left[y_1, \cdots, y_T\right]$ to denote the general input and output sequences without referring to specific sequences. Therefore,  for $T$ random variables $y_1, \cdots, y_T$ i.i.d sampled from a distribution $F_Y$, the distribution of the maximum is realised using EVT as :
\begin{equation}
\lim _{T \rightarrow \infty} P\left\{\max \left(y_1, \cdots, y_T\right) \leq y\right\}=\lim _{T \rightarrow \infty} F^T(y)=G(y),
\label{Gev_def_paper}
\end{equation}
for some linear transformation where $G(y)$ is GEV distribution function. We can observe that equation (\ref{Gev_def}) and equation (\ref{Gev_def_paper}) have the same meaning (but with different variables in their definitions). Moreover, the authors define the GEV function as:
\begin{equation}
G(y)=\left\{\begin{array}{cc}
\exp \left(-\left(1-\frac{1}{\gamma} y\right)^\gamma\right), &  \gamma \neq 0,1-\frac{1}{\gamma} y>0 \\
\exp \left(-e^{-y}\right), &  \gamma=0,
\end{array}\right.
\label{GEV function paper}
\end{equation}
where $\gamma$ is known as the extreme value index (the shape parameter) with condition $\gamma \neq 0$. It can also be observed that the definition of GEV function in equation (\ref{GEV function paper}) is similar to the definition mentioned in equation (\ref{GEV function}) but with $\xi = - \frac{1}{\gamma}$, $\mu = 0$ and, $\sigma = 1$. For modeling the tail distribution of the corresponding time-series data, they use equation (\ref{DeHaan Tail approx.}) but as mentioned before, rather than using the GPD function they use the GEV distribution function to model the tail approximation. Therefore, we substitute the relationship mentioned in equation (\ref{Rel. GPD & GEV}) as $-ln(G(y)) = 1 - H(y)$ in equation (\ref{DeHaan Tail approx.}) and get the following result:
\begin{equation}
1-F(y) \approx(1-F(\xi))\left[-\ln G\left(\frac{y-\xi}{f(\xi)}\right)\right], y>\xi,
\label{tail approx. paper}
\end{equation}
where $\xi$ is the threshold and, $f(\xi)$ is a scale function as mentioned in the paper \cite{Haroan}. Also, the authors define an extreme indicator sequence $V_{1: T} = \left[v_1, \cdots, v_T\right]$ as:
\begin{equation}
v_t=\left\{\begin{array}{cc}
1 & y_t>\xi \\
0 & y_t\leqslant\xi,
\end{array}\right.
\label{extrem indicator}
\end{equation}
where $\xi$ is the threshold. For time step $t$ if $v_t = 0$ then the output $y_t$ is considered as a 'normal event' and if $v_t = 1$ then $y_t$ is considered as an 'extreme event'. The authors also mention a hard approximation for the term $(\frac{y-\xi}{f(\xi)})$ in equation (\ref{tail approx. paper}) as $u_t$ which is the predicted indicator by the neural network used by them in their experiment. This can be interpreted as a normalization which restricts the values of output $y$, above and below the threshold $\xi$ between $[-1, 1]$. Therefore, considering this to be true, we can rewrite equation (\ref{tail approx. paper}) as:
\begin{equation}
1-F(y) \approx(1-F(\xi))\left[-\ln G(u_t)\right],
\label{tail approx. paper - 1}
\end{equation}
Substituting the definition of GEV in equation (\ref{GEV function paper}) into the above equation (\ref{tail approx. paper - 1}) we obtain:
\begin{equation}
1-F(y) \approx(1-F(\xi))\left[1-\frac{u_t}{\gamma}\right]^\gamma.
\label{tail approx. paper - 2}
\end{equation}
The term $1 - F(\xi)$ can be approximated as:
\begin{equation}
\begin{aligned}
1-F(\xi) 
= & \operatorname{Pr}(y>\xi) 
\implies 1-F(\xi) = \operatorname{Pr}(v_t=1),
\end{aligned}
\end{equation}
where $\operatorname{Pr}(v_t=1)$ is the proportion of extreme events in the dataset.
Therefore, we can rewrite equation (\ref{tail approx. paper - 2}) with the above substitution as:
\begin{equation}
1-F(y) \approx \operatorname{Pr}(v_t=1) \left[1-\frac{u_t}{\gamma}\right]^\gamma.
\label{tail approx. mine}
\end{equation}
This tail approximation is incorporated in the terms of the standard Cross Entropy (CE) function as weights to define the main EVL loss function mentioned in paper \cite{Haroan}. However, the authors in paper \cite{Haroan} define the weight as:
\begin{equation}
1-F(y) \approx (1 - \operatorname{Pr}(v_t=1)) \left[1-\frac{u_t}{\gamma}\right]^\gamma.
\label{tail approx. paper - 3}
\end{equation}
Upon simplifying the term $(1 - \operatorname{Pr}(v_t=1)$ we get:
\begin{equation}
\begin{aligned}
& 1-\operatorname{Pr}\left(v_t=1\right) \\
= & \operatorname{Pr}\left(v_t=0\right) \\
= & \operatorname{Pr}(y \leqslant \xi) \\
= & F(\xi),
\end{aligned}
\end{equation}
so we get the expression $1-F(y) \approx(F(\xi))\left[1-\frac{u_t}{\gamma}\right]^\gamma$ which is not in congruence with the main tail approximation in equation (\ref{tail approx. paper}) as shown by \cite{Haroan}. Moreover, research by \cite{Chen2023-iu} show similar weight derivations for the EVL loss function as it has been derived in equation (\ref{tail approx. mine}). Therefore, applying the weights derived in equation (\ref{tail approx. mine}) to the standard BCE loss function, we get:

\begin{equation}
\begin{aligned}
\operatorname{EVL}\left(u_t,v_t\right)
= & -\operatorname{Pr}(v_t=1)\left[1-\frac{u_t}{\gamma}\right]^\gamma v_t \log \left(u_t\right) \\
& -\operatorname{Pr}(v_t=0)\left[1-\frac{1-u_t}{\gamma}\right]^\gamma\left(1-v_t\right) \log \left(1-u_t\right),
\label{EVL loss function}
\end{aligned}
\end{equation}
where the standard BCE loss function for a binary classification task is given by:
\begin{equation}
\begin{aligned}
\operatorname{BCE}\left(u_t,v_t\right)
= & -v_t \log \left(u_t\right) \\
& -\left(1-v_t\right) \log \left(1-u_t\right).
\label{BCE loss function}
\end{aligned}
\end{equation}
\newpage

\subsection{Additional Results}
\label{appendix:results}

\begin{figure}[ht]
    \centering
    \includegraphics[width=15cm]{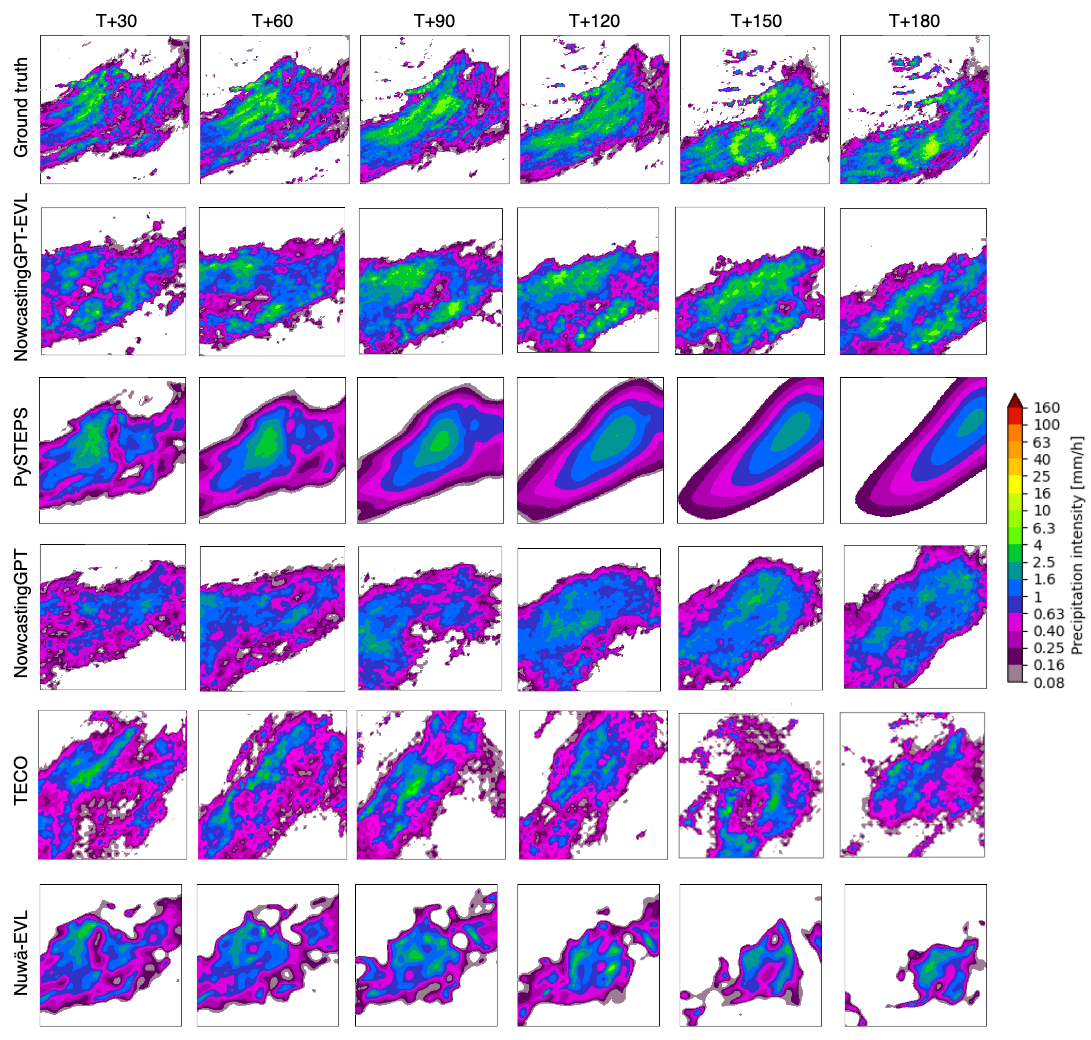}
    \caption{Nowcasting of extreme precipitation scenarios. The generation is conditioned on $3$ previous timestamps with the task to predict the next $6$ lead times. There is a gap of $30$ minutes between each timestamp. Images are upsampled to $256\times256$ pixels.}
    \label{img:reconstructions}
\end{figure}

\begin{figure}[ht]
    \centering
    \includegraphics[width=14cm]{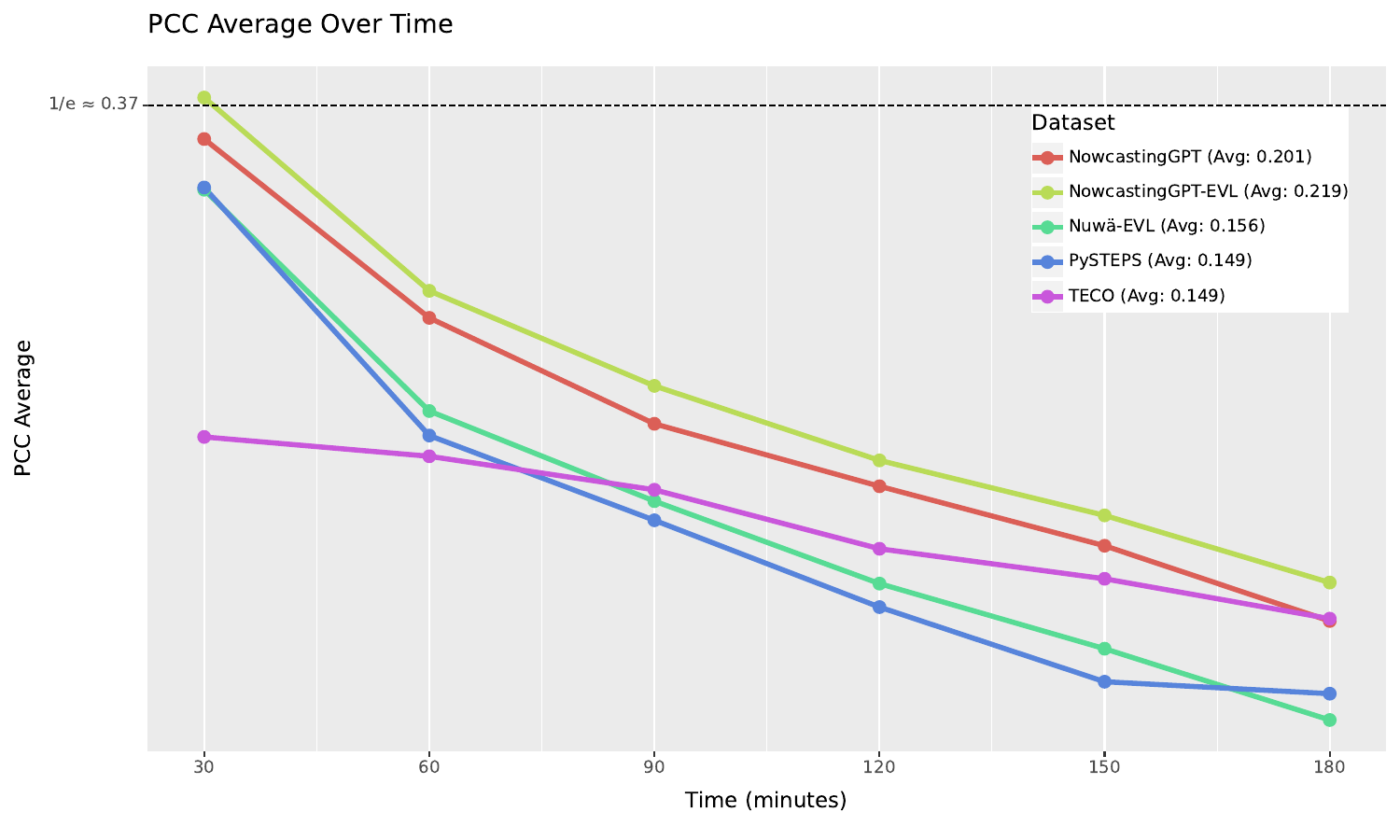}
    \caption{PCC metric evaluation over the 6 lead times. Each point represents the average value for a specific lead time over the whole dataset. Higher values represent better performance. NowcastingGPT-EVL outperforms all other models.}
    \label{img:fig3}
\end{figure}

\begin{figure}[ht]
    \centering
    \includegraphics[width=14cm]{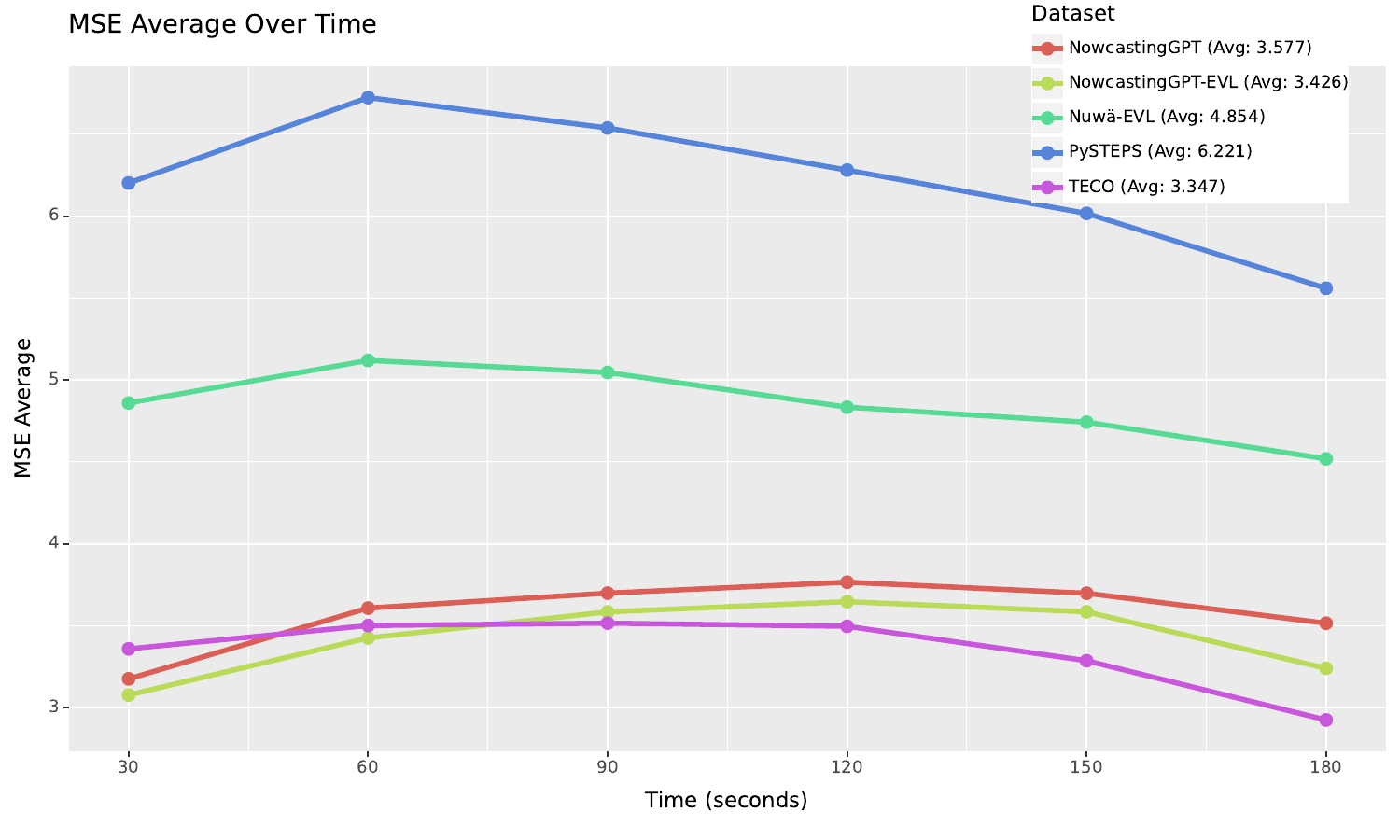}
    \caption{MSE metric evaluation over the 6 lead times. Each point represents the average value for a specific lead time over the whole dataset. Lower values represent better performance. NowcastingGPT-EVL and TECO outperforms all other models for bigger lead times.}
    \label{img:overlay1}
\end{figure}

\begin{figure}[ht]
    \centering
    \includegraphics[width=14cm]{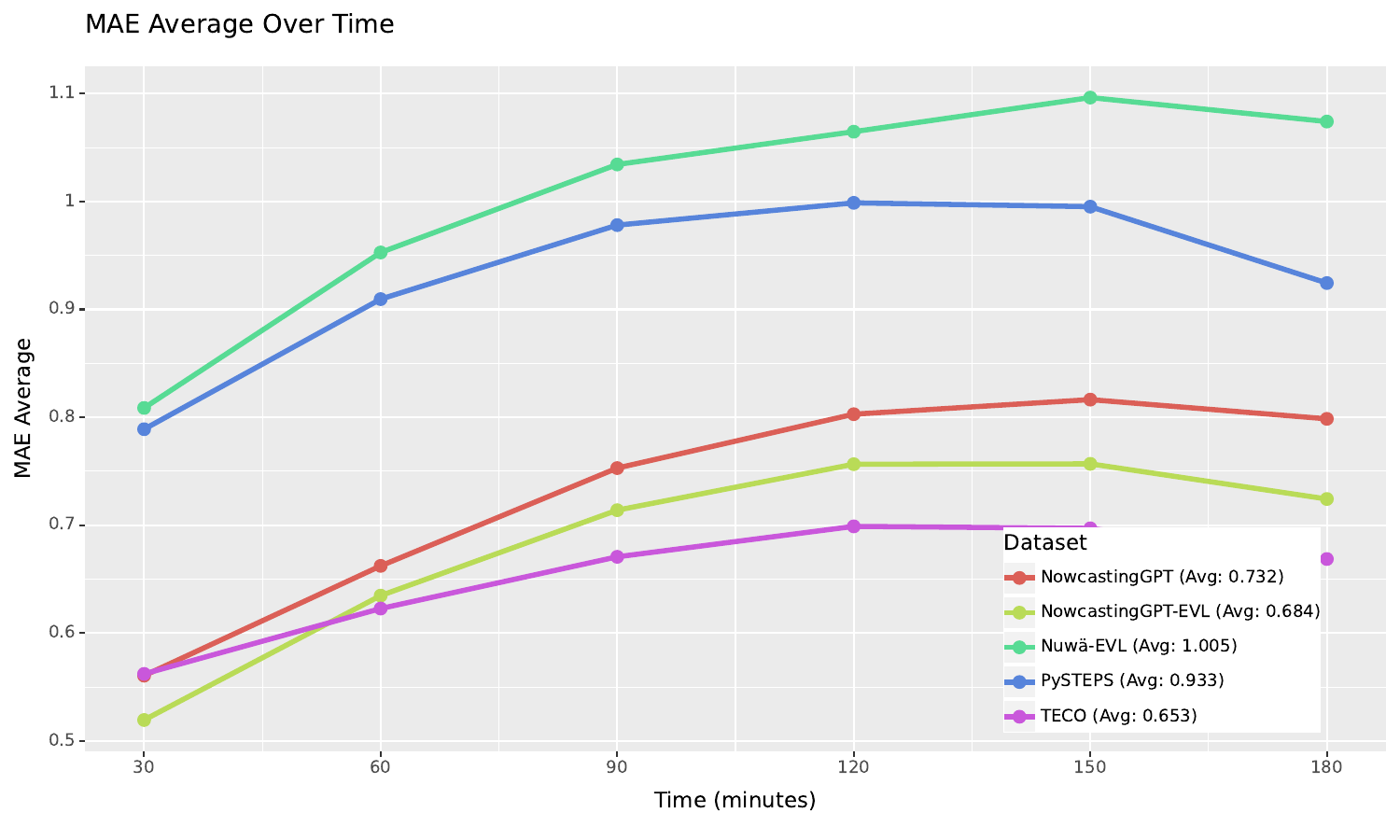}
    \caption{MAE metric evaluation over the 6 lead times. Each point represents the average value for a specific lead time over the whole dataset. Lower values represent better performance. TECO outperforms all other models.}
    \label{img:overlay2}
\end{figure}

\begin{figure}[ht]
    \centering
    \includegraphics[width=14cm]{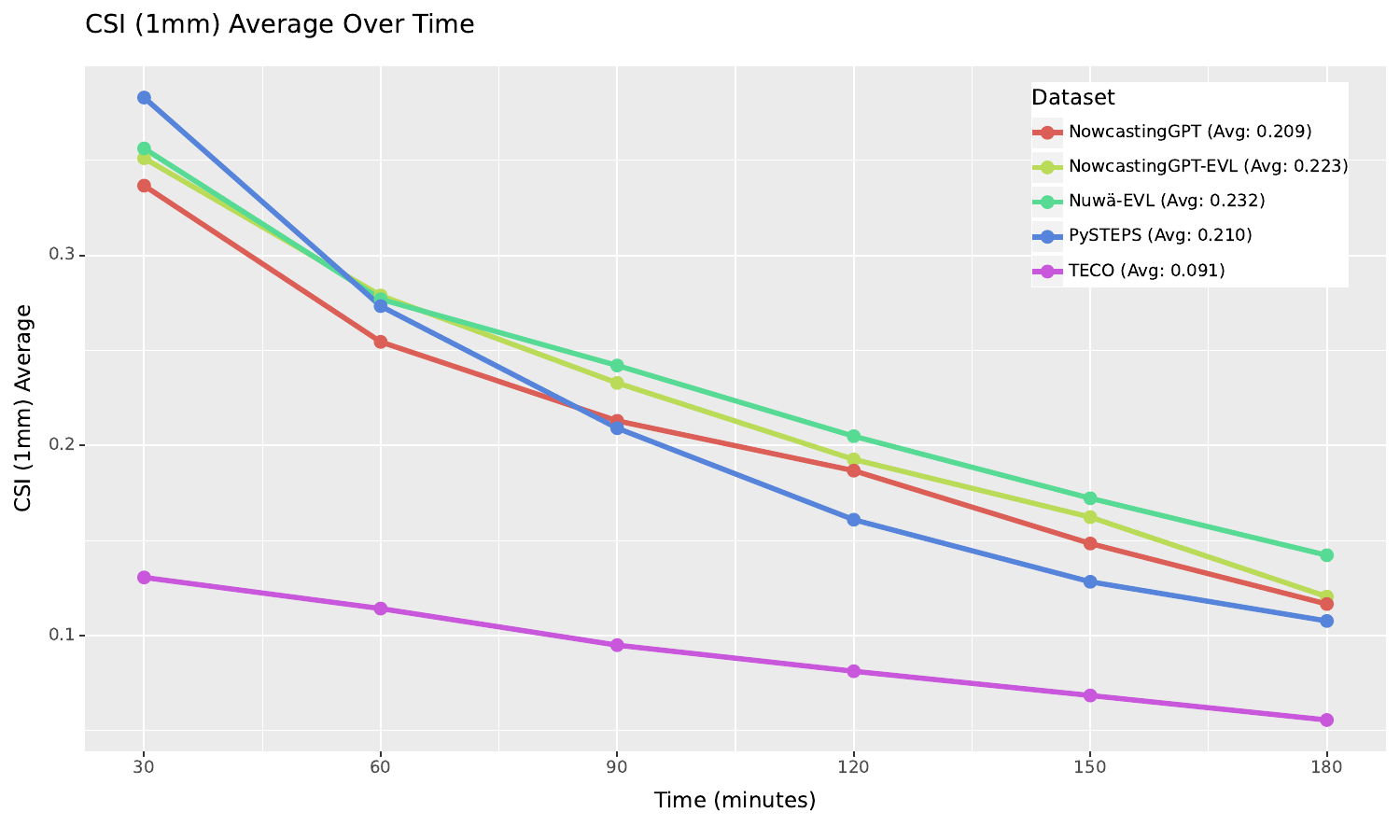}
    \caption{CSI(1mm) metric evaluation over the 6 lead times. Each point represents the average value for a specific lead time over the whole dataset. Higher values represent better performance. Nuwä-EVL and NowcastingGPT-EVL outperform the rest of the models but decay quickly.}
    \label{img:overlay3}
\end{figure}

\begin{figure}[ht]
    \centering
    \includegraphics[width=14cm]{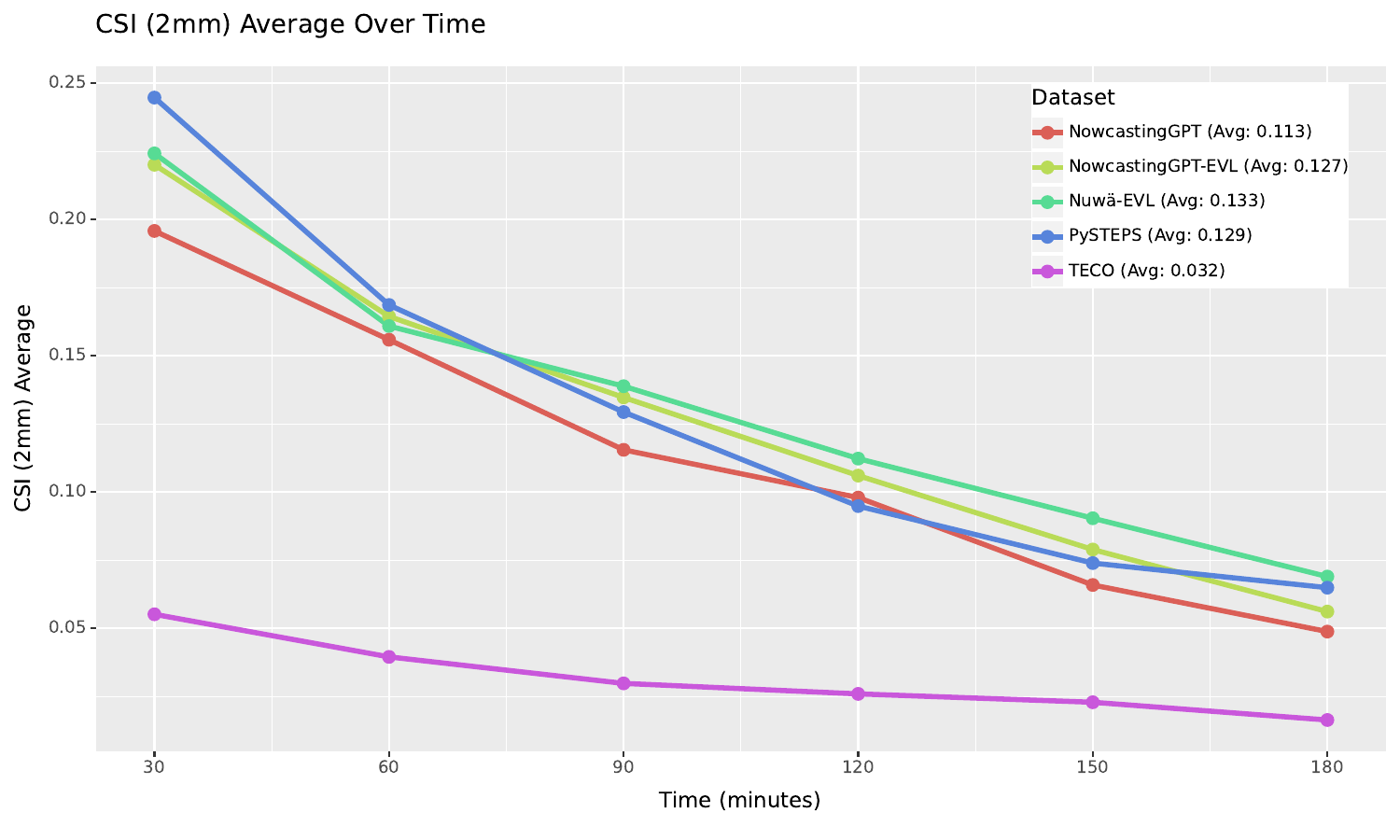}
    \caption{CSI(2mm) metric evaluation over the 6 lead times. Each point represents the average value for a specific lead time over the whole dataset. Higher values represent better performance. Nuwä-EVL outperforms the rest of the models.}
    \label{img:overlay4}
\end{figure}

\begin{figure}[ht]
    \centering
    \includegraphics[width=14cm]{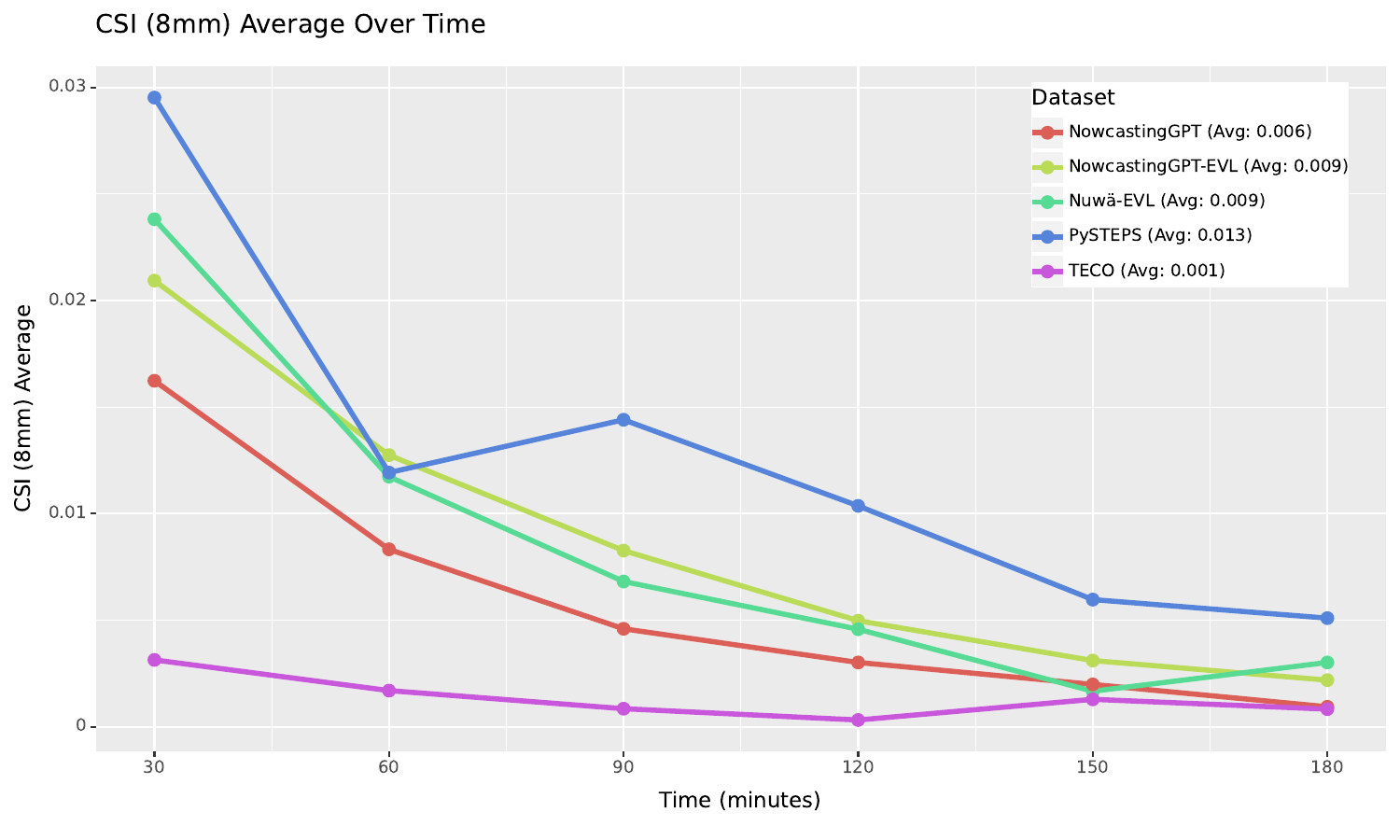}
    \caption{CSI(8mm) metric evaluation over the 6 lead times. Each point represents the average value for a specific lead time over the whole dataset. Higher values represent better performance. PySTEPS outperforms the rest of the models.}
    \label{img:overlay5}
\end{figure}

\begin{figure}[ht]
    \centering
    \includegraphics[width=14cm]{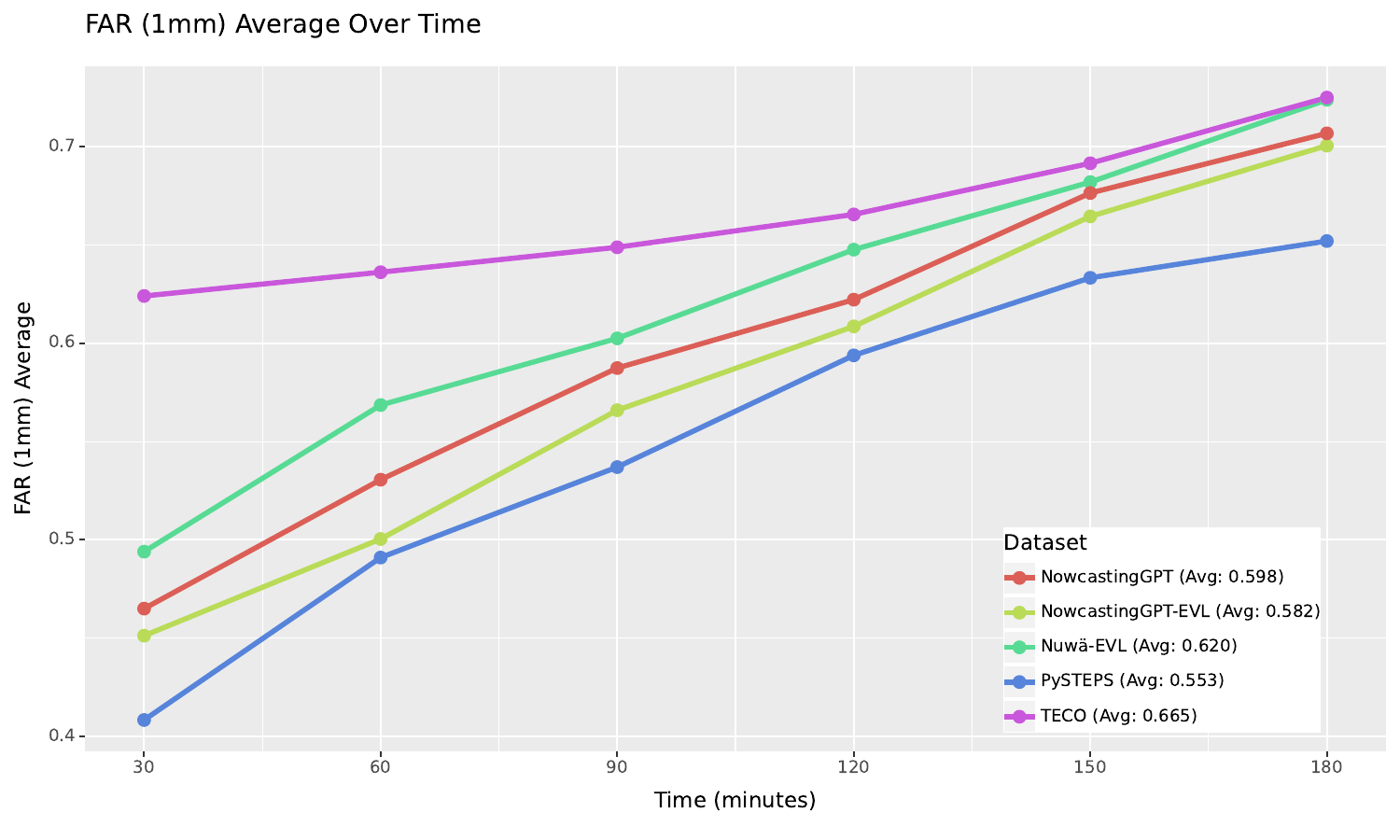}
    \caption{FAR(1mm) metric evaluation over the 6 lead times. Each point represents the average value for a specific lead time over the whole dataset. Lower values represent better performance. PySTEPS outperforms the rest of the models.}
    \label{img:overlay6}
\end{figure}

\begin{figure}[ht]
    \centering
    \includegraphics[width=14cm]{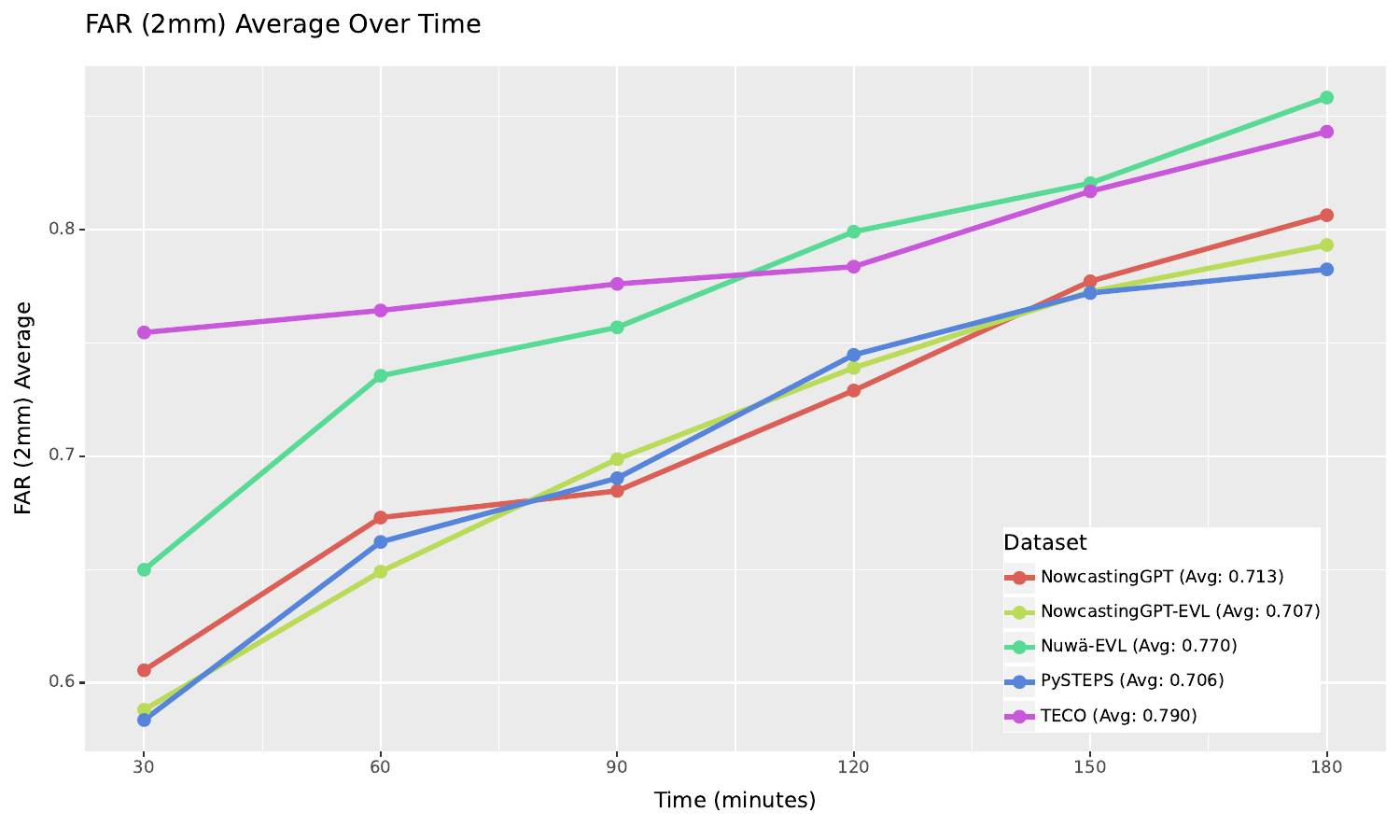}
    \caption{FAR(2mm) metric evaluation over the 6 lead times. Each point represents the average value for a specific lead time over the whole dataset. Lower values represent better performance. NowcastingGPT-EVL outperforms the rest of the models.}
    \label{img:overlay7}
\end{figure}

\begin{figure}[ht]
    \centering
    \includegraphics[width=14cm]{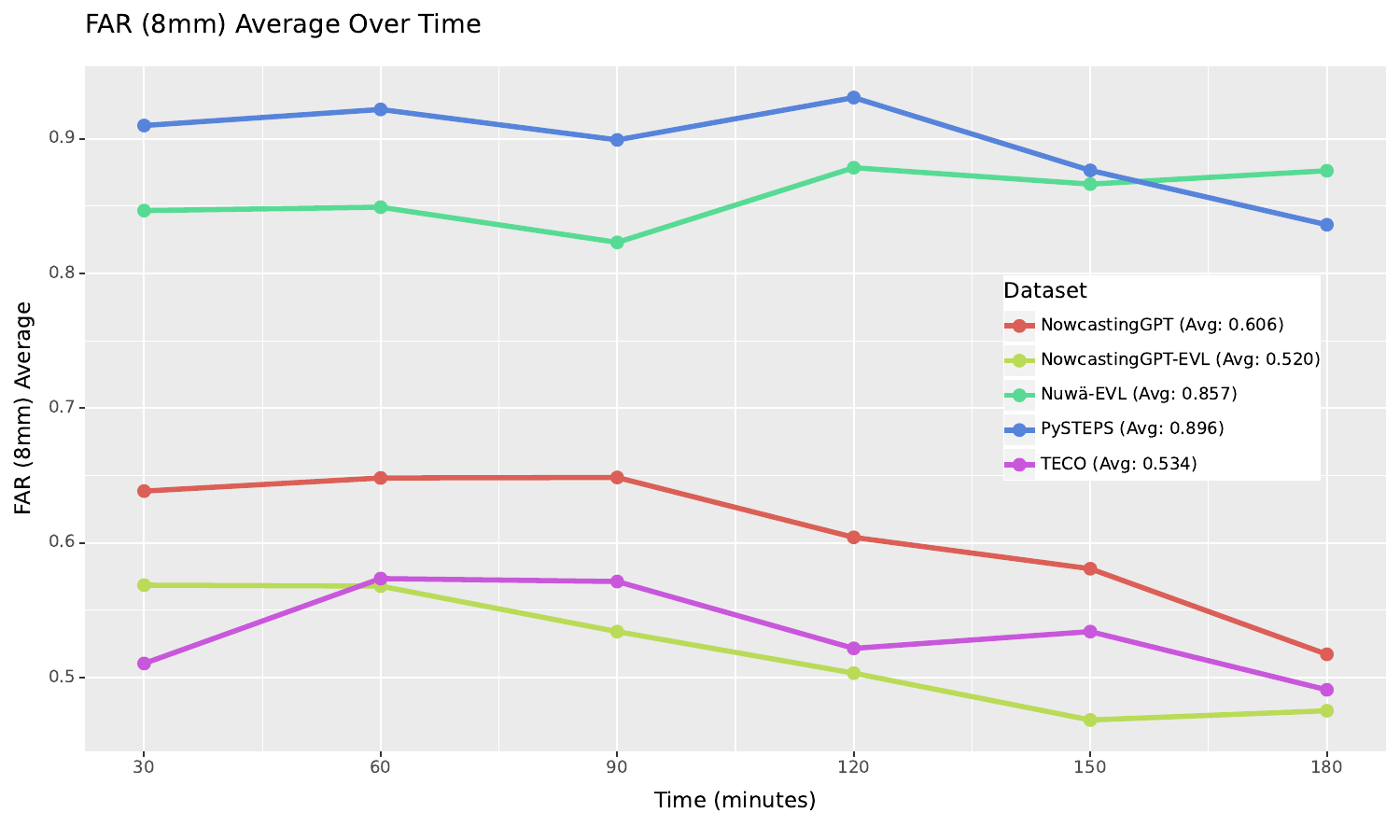}
    \caption{FAR(8mm) metric evaluation over the 6 lead times. Each point represents the average value for a specific lead time over the whole dataset. Lower values represent better performance. NowcastingGPT-EVL and TECO outperform the rest of the models.}
    \label{img:overlay8}
\end{figure}

\begin{figure}[ht]
    \centering
    \includegraphics[width=14cm]{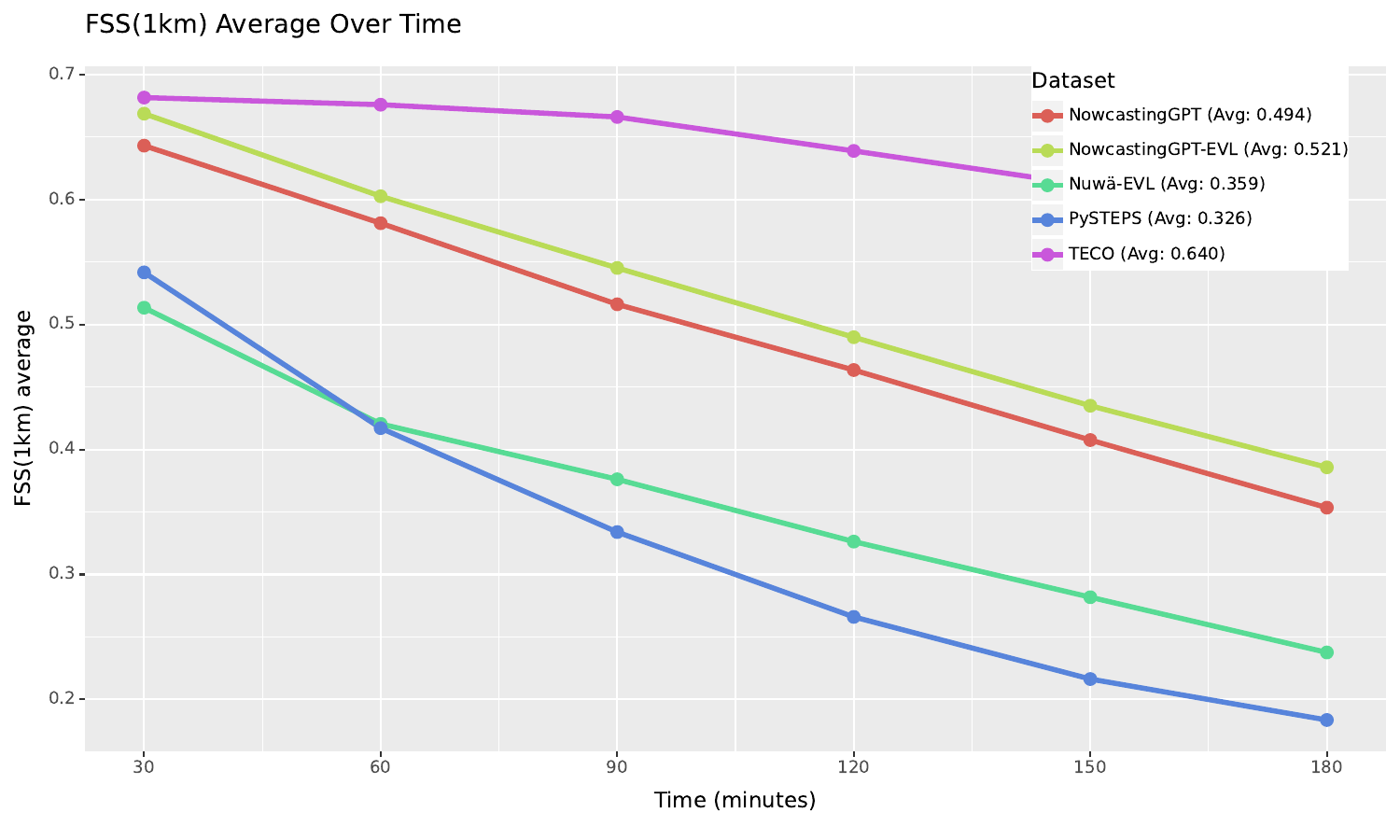}
    \caption{FSS(1km) metric evaluation over the 6 lead times. Each point represents the average value for a specific lead time over the whole dataset. Higher values represent better performance. TECO outperforms the rest of the models.}
    \label{img:overlay9}
\end{figure}

\begin{figure}[ht]
    \centering
    \includegraphics[width=14cm]{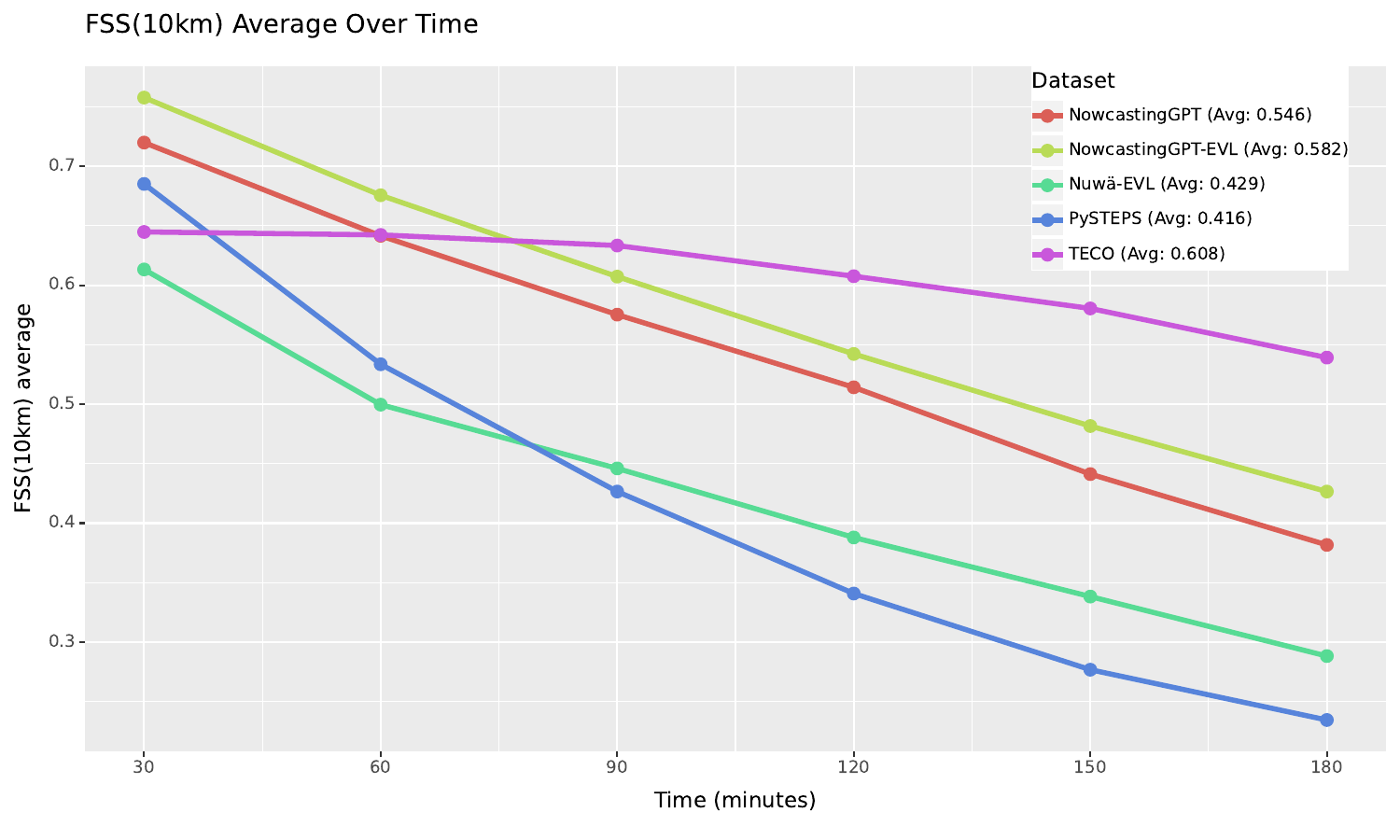}
    \caption{FSS(10km) metric evaluation over the 6 lead times. Each point represents the average value for a specific lead time over the whole dataset. Higher values represent better performance. TECO outperforms the rest of the models on higher lead times while NowcastingGPT and NowcastingGPT-EVL perform better on lower lead times.}
    \label{img:overlay10}
\end{figure}

\begin{figure}[ht]
    \centering
    \includegraphics[width=14cm]{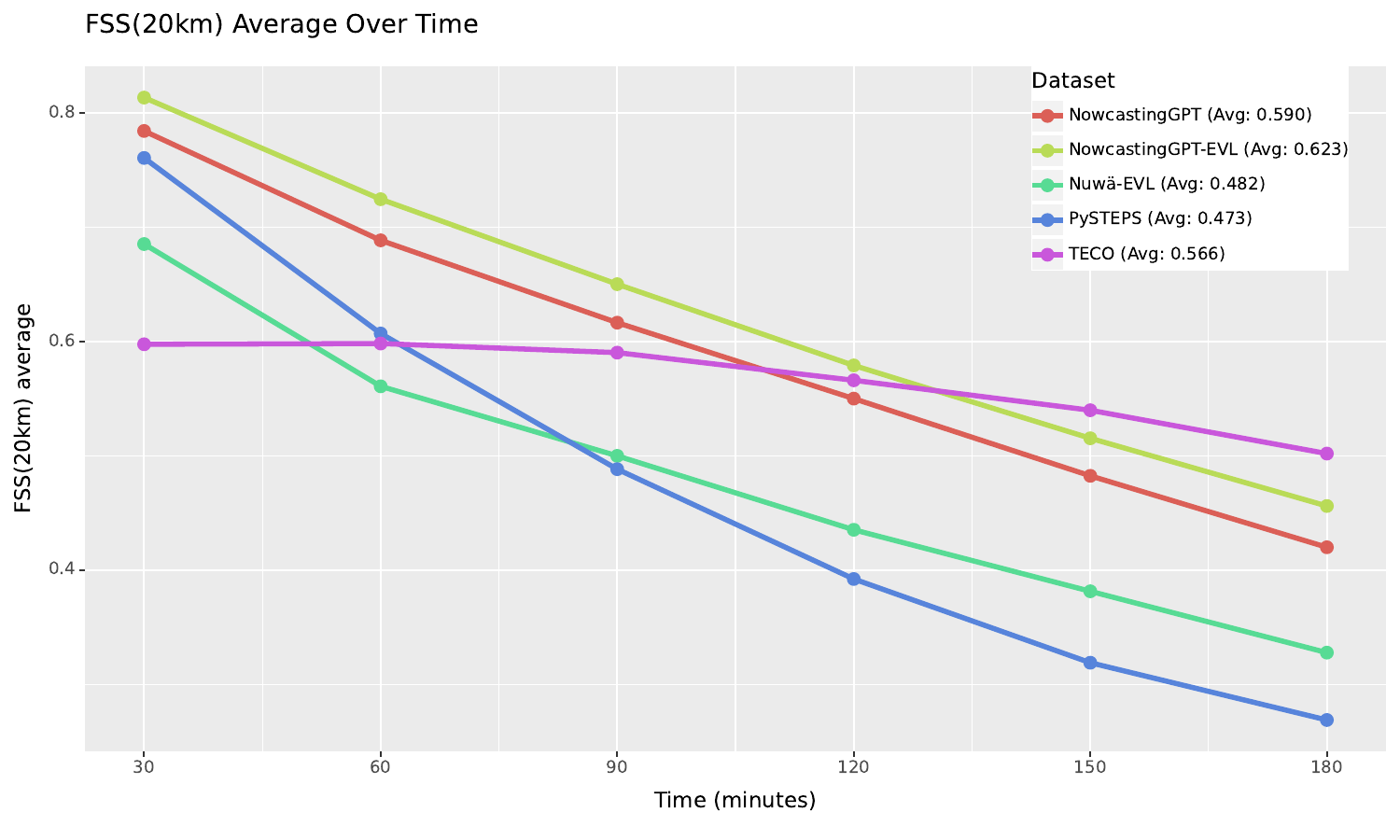}
    \caption{FSS(20km) metric evaluation over the 6 lead times. Each point represents the average value for a specific lead time over the whole dataset. Higher values represent better performance. NowcastingGPT-EVL outperforms the rest of the models.}
    \label{img:overlay11}
\end{figure}

\begin{figure}[ht]
    \centering
    \includegraphics[width=14cm]{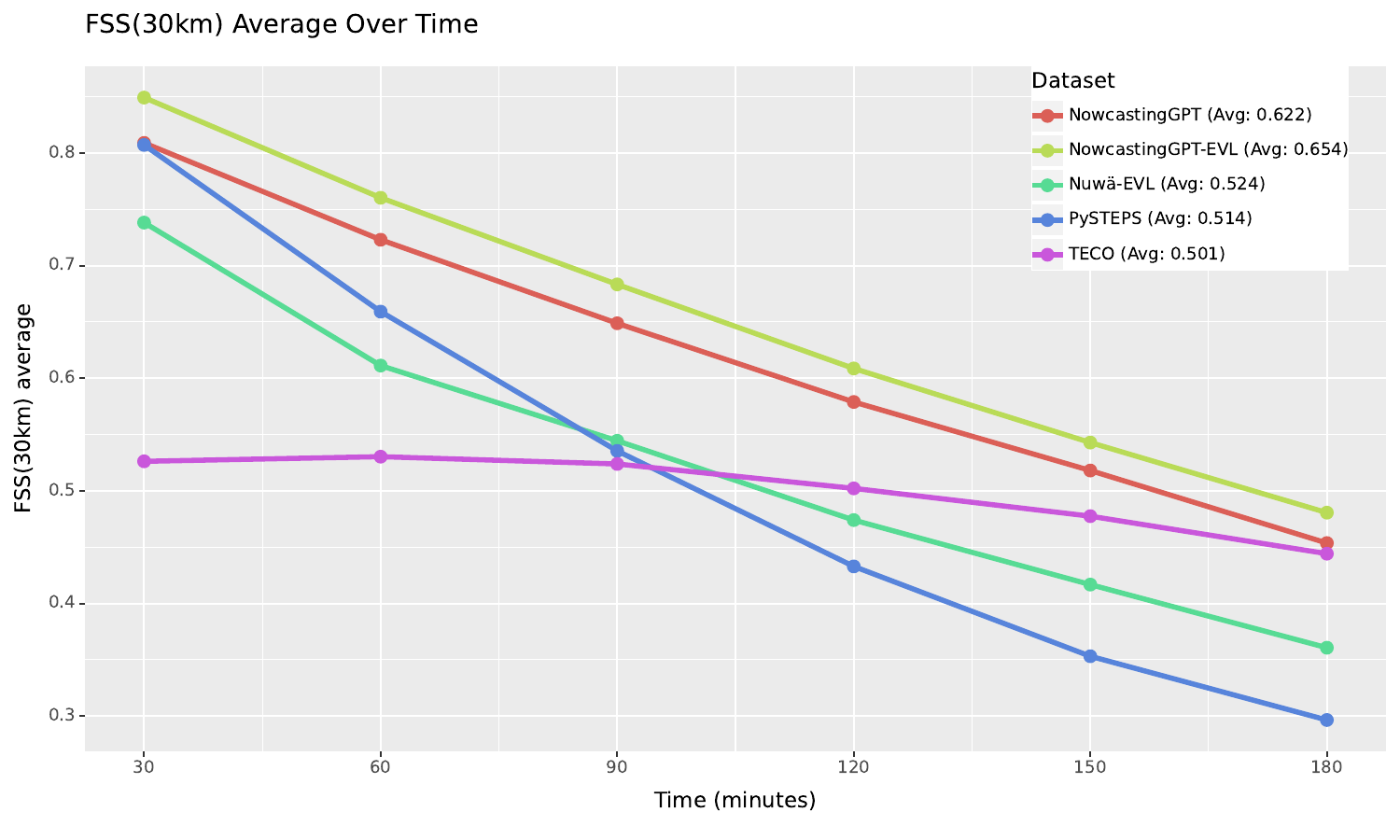}
    \caption{FSS(30km) metric evaluation over the 6 lead times. Each point represents the average value for a specific lead time over the whole dataset. Higher values represent better performance. NowcastingGPT-EVL outperforms the rest of the models.}
    \label{img:last}
\end{figure}

This section supports the findings in Section \ref{experimental} with a series of qualitative results that provide a different perspective for assessing the quality of the results. Thus, the emphasis is on the quality difference between lead times on the extreme precipitation events dataset. Figure \ref{img:reconstructions} aims to provide visualizations of the predicted lead times as a visual signal on the quality of the predictions. On the other hand, Figure \ref{img:fig3} to Figure \ref{img:last} provide additional analysis on the performance of the proposed models on the metrics presented in Section \ref{experimental}


\end{document}